\pdfoutput=1
\documentclass{article}

\PassOptionsToPackage{numbers, compress}{natbib}

\usepackage[main, final]{neurips_2026}

\usepackage[utf8]{inputenc}
\usepackage[T1]{fontenc}
\usepackage{hyperref}
\usepackage{url}
\usepackage{booktabs}
\usepackage{colortbl}
\usepackage{amsfonts}
\usepackage{microtype}
\usepackage[dvipsnames]{xcolor}

\usepackage{graphicx}
\graphicspath{{figures/}}

\usepackage{amsmath}
\usepackage{amssymb}
\usepackage{mathtools}
\usepackage{amsthm}

\usepackage[capitalize,noabbrev]{cleveref}

\theoremstyle{plain}
\newtheorem{theorem}{Theorem}[section]

\newtheorem{lemma}[theorem]{Lemma}

\theoremstyle{definition}

\newtheorem{assumption}[theorem]{Assumption}
\theoremstyle{remark}

\usepackage{subcaption}
\usepackage{tikz}

\usepackage{algorithm}
\usepackage{algpseudocode}

\usepackage[most]{tcolorbox}

\algrenewcommand\algorithmicindent{1.1em}
\algrenewcommand\algorithmiccomment[1]{\hfill{\footnotesize\color{gray}$\triangleright$~#1}}

\Crefname{section}{Section}{Sections}
\Crefname{table}{Table}{Tables}
\Crefname{figure}{Figure}{Figures}
\Crefname{assumption}{Assumption}{Assumptions}

\hypersetup{
    colorlinks = true,
    linkcolor = red,
    anchorcolor = red,
    urlcolor = red,
    citecolor = blue
}

\usepackage{xspace}
\makeatletter
\DeclareRobustCommand\onedot{\futurelet\@let@token\@onedot}
\def\@onedot{\ifx\@let@token.\else.\null\fi\xspace}
 
\def\ie{\emph{i.e}\onedot} 
 
 \def\vs{\emph{vs}\onedot}

\makeatother

\tcbset{
  myeqbox/.style={
    enhanced,
    colback=gray!10,
    colframe=gray,
    boxrule=1pt,
    arc=2pt,
    left=20pt,
    right=20pt,
    top=8pt,
    bottom=8pt,
    boxsep=0pt
  }
}

\usepackage{amsmath, amsfonts, amsthm, bm}

\def\eqref#1{equation~\ref{#1}}

\def\1{\bm{1}}

\def\ve{{\bm{e}}}

\def\vg{{\bm{g}}}

\def\vs{{\bm{s}}}

\def\vw{{\bm{w}}}

\DeclareMathAlphabet{\mathsfit}{\encodingdefault}{\sfdefault}{m}{sl}
\SetMathAlphabet{\mathsfit}{bold}{\encodingdefault}{\sfdefault}{bx}{n}

\newcommand{\R}{\mathbb{R}}

\def\R{\mathbb{R}}

\definecolor{myred}{RGB}{255,133,133}
\definecolor{mygrey}{RGB}{104,104,104}
\definecolor{paramred}{RGB}{215,17,41}
\newcommand{\blackdiamond}{\rotatebox[origin=c]{45}{$\vcenter{\hbox{$\scriptscriptstyle\blacksquare$}}$}}

\newcommand{\ours}{\textsc{Fresco}\xspace}

\title{\ours: A Novel Consistency Control\\ for Asynchronous Pipeline Parallel Training}

\author{%
    Dongyeop Lee \\
    POSTECH\\
    \texttt{dongyeop.lee2@postech.ac.kr}\\
    \And
    Namhoon Lee\\
    POSTECH\\
    \texttt{namhoon.lee@postech.ac.kr} \\
}

\begin{document}

\maketitle

\begin{abstract}
Asynchronous pipeline parallelism can accelerate distributed model training, but suffers from instability caused by inconsistent gradient computations.
We identify a fundamental source of this inconsistency: error signals are coupled to forward-time features, not just parameter states.
This means that parameters may evolve as long as they preserve \emph{forward-equivalence}.
Leveraging this insight, we introduce \ours, which frames asynchronous updates as a constrained optimization problem: it finds the minimal update modification that acts trivially on the activation subspace.
We demonstrate that \ours{} consistently outperforms representative asynchronous baselines in LLM pre-training, with the gains most pronounced under deep pipelines and high inconsistency levels: at pipeline depth $32$, it reaches the same loss as the strongest baseline with $43\%$ fewer iterations and $30\%$ less wall-clock time, while requiring $8$--$921\times$ less memory overhead.
Moving beyond the current limitations, \ours{} establishes a new foundation for high-utilization pipeline parallel training, providing a scalable path for large-model development on diverse, resource-constrained systems.
\end{abstract}

\section{Introduction}
\label{sec:intro}

Pipeline parallelism has played a central role in the scaling of modern machine learning by partitioning enormous neural networks across multiple processing units along their sequential data-processing structure, enabling large-scale distributed training across hundreds or even thousands of devices \citep{dubey2024llama,liu2024deepseek,team2025kimi}.
Pipeline scheduling has been integral to fully exploiting the computational capabilities of these systems, continuously streaming micro-batches through successive stages to keep different processing units concurrently occupied rather than idle \citep{petrowski1993performance,kosson2021pipelined,gaunt2017ampnet,huo2018decoupled,huang2019gpipe,narayanan2019pipedream}.

\begin{figure}
    \centering
    \begin{subfigure}{0.49\linewidth}
        \centering
        \includegraphics[width=\linewidth,trim={0.8em 0em -1em 0em},clip]{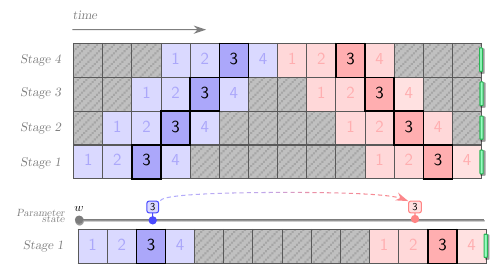}
        \caption{Synchronous pipeline parallelism}
        \label{fig:async-gantt-a}
    \end{subfigure}
    \hfill
    \begin{subfigure}{0.49\linewidth}
        \centering
        \includegraphics[width=\linewidth,trim={0em 0em 1em 0em},clip]{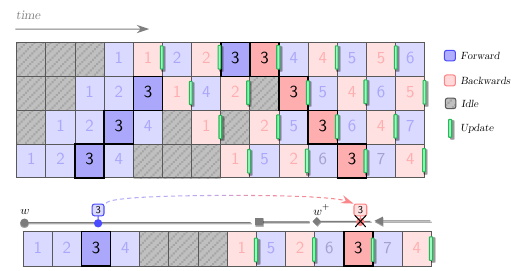}
        \caption{Asynchronous pipeline parallelism}
        \label{fig:async-gantt-b}
    \end{subfigure}
\caption{Gantt charts illustrating stage-processor states over time for micro-batch pipelining under synchronous and asynchronous parameter updates. Asynchronous updating allows the stage-1 threads for micro-batches 1 and 2 to overwrite the parameters while micro-batch 3 is still waiting on its downstream error signal \textcolor{myred}{$\bullet$} after reading the previous parameter state \textcolor{mygrey}{$\bullet$}, thus the updated parameters \textcolor{mygrey}{$\blackdiamond$} no longer form a proper derivative chain, degrading its utility for improving stage 1.}
    \label{fig:async-gantt}
\end{figure}

Asynchronous pipelining offers a straightforward way to achieve sustained, bubble- and halt-free pipeline execution by allowing parameter updates to proceed concurrently with the forward and backward computations of in-flight micro-batches (see \cref{fig:async-gantt} for illustrations).
However, lifting this parameter-equivalence condition can introduce two types of inconsistencies into these concurrent gradient computations: \emph{chain inconsistency}, where partial gradients are chained across different parameter versions, and \emph{gradient staleness}, where even a chain-consistent gradient no longer corresponds to the current parameter state, both of which can degrade training stability and convergence.
To mitigate these adverse effects, various correction mechanisms have been developed to compute gradients with respect to either past parameter versions, obtained through stashing or reconstruction \citep{narayanan2019pipedream,narayanan2021memory,yang2021pipemare}, or predicted future parameter states \citep{chen2018efficient,guan2019xpipe,guan2025pipeoptim}.

Yet, prior methods have largely overlooked the possibility of controlling parameter updates themselves to preserve concurrent gradient computations consistent, even though preventing interference with these computations was the very motivation for imposing the restriction in the first place.
Consequently, asynchronous pipeline optimization has yet to establish what principle should replace the parameter-equivalence condition that safeguards gradient computations in synchronous execution.
Instead, existing methods have primarily focused on compensating post hoc for errors within individual micro-batches, rather than controlling the inter-micro-batch interference that gives rise to them.
Simply lifting the restriction to permit unconstrained parameter updates provides no principled criterion for determining which changes can preserve gradient consistency without introducing staleness.

In this work, we address this gap by showing that parameter changes degrade gradient computation not simply by altering the parameters, but by breaking the compatibility between a returned error signal and the forward behavior of the updated stage at the inter-stage boundary.
Concretely, when the updated parameters would produce a different stage output from the one on which the error signal was evaluated, the error signal and the local Jacobian no longer form a valid gradient for the updated parameters.
This reveals that the relevant consistency requirement is preserving the forward behavior seen by in-flight micro-batches even if the full parameter tensor changes, \ie, \emph{forward-equivalence} over parameter-equivalence.

\begin{figure}[t]
    \centering
    \begin{subfigure}[t]{0.31\linewidth}
        \centering
        \includegraphics[width=\linewidth,trim={0 0em 0 0},clip]{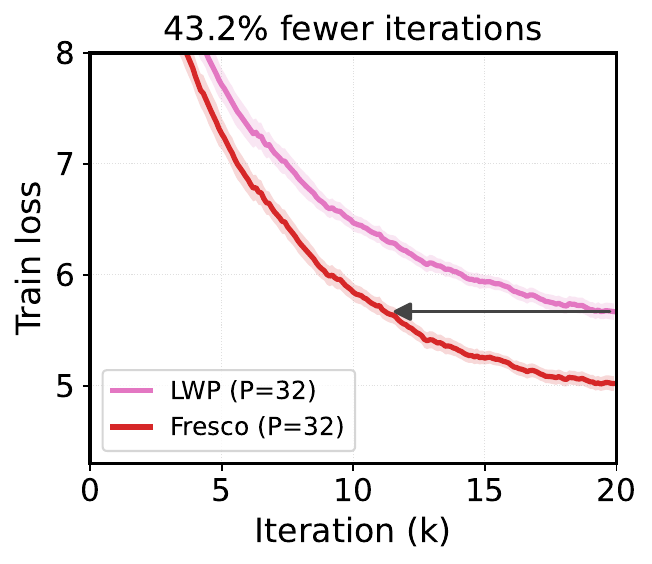}
        \caption{Robust under deep pipeline}
        \label{fig:intro-effectiveness}
    \end{subfigure}
    \hfill
    \begin{subfigure}[t]{0.32\linewidth}
        \centering
        \includegraphics[width=\linewidth,trim={0 0.6em 0 0},clip]{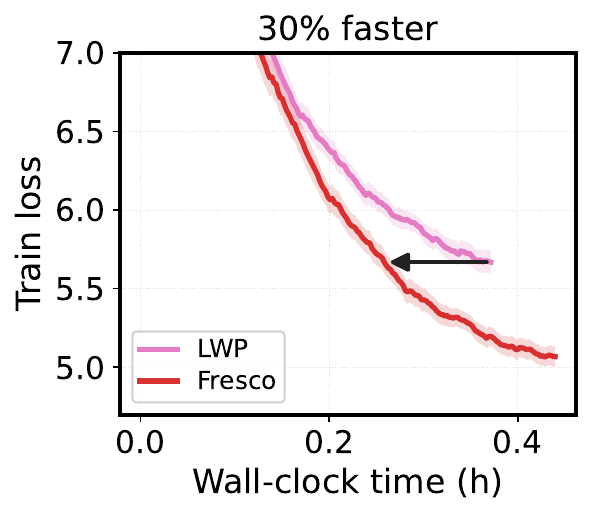}
        \caption{Fast training}
        \label{fig:intro-fast}
    \end{subfigure}
    \hfill
    \begin{subfigure}[t]{0.34\linewidth}
        \centering
        \includegraphics[width=\linewidth,trim={0 0.5em 0 0},clip]{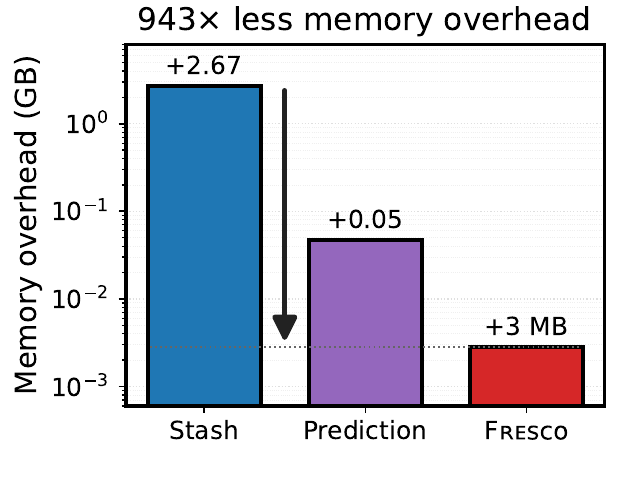}
        \caption{Memory efficiency}
        \label{fig:intro-memory}
    \end{subfigure}
    \caption{
    Summary of results.
    (a) Under a deep asynchronous pipeline, \ours improves optimization over best baseline, reaching the same training loss with fewer iterations, (b) yielding faster training speeds under wall-clock time.
    (c) \ours avoids the large memory overhead of maintaining multiple full parameter versions, using substantially less memory than stashing-based approaches.
    }
    \label{fig:intro-summary}
    \vspace{-0.5em}
\end{figure}

Guided by this insight, we propose \emph{\ours}, an asynchronous pipeline-parallel training method that controls parameter updates according to their effect on the forward activations of in-flight micro-batches.
Concretely, we formulate this as a constrained optimization problem that finds the closest modification of each optimizer update subject to acting trivially on the activation subspace occupied by in-flight micro-batches.
To implement this efficiently, \ours maintains a compact spectral basis for this activation subspace and projects each layer-wise update onto its orthogonal complement, thereby removing exactly the components that would violate forward equivalence.
This allows parameters to evolve asynchronously while preserving the forward behavior required for the returned error signals to remain valid.
Finally, we establish convergence of \ours{}, showing that it control the degree of perturbation introduced during forward and backwards pass to guarantee convergence.

Empirically, \ours{} improves asynchronous pipeline-parallel training across model scales and pipeline depths, achieving lower loss than representative asynchronous baselines in an established pipeline-parallel pre-training setting with a 95M-parameter language model, especially at moderate and deep pipeline depths (\cref{fig:intro-effectiveness}), allowing \ours{} to achieve the same loss with 30\% less wall-clock time (\cref{fig:intro-fast}).
Moreover, because \ours caches only compact dominant feature spectra rather than full parameter copies, its memory overhead remains marginal compared with existing baselines, especially for larger models (\cref{fig:intro-memory}).

These results indicate that \ours provides an effective foundation for reliable, high-utilization pipeline-parallel training, opening a new avenue beyond strict parameter consistency toward scalable training on conventional and resource-constrained collaborative systems.

\section{Problem analysis}
\label{sec:background}

We begin by presenting a detailed formulation of the parameter inconsistency problem in asynchronous pipeline-parallel training and the forward-equivalence condition.

\paragraph{Preliminary}
A pipeline-parallel training setup partitions a model $F = f_P \circ \cdots \circ f_1$ across multiple processing units, each responsible for a stage $f_p$.
Each stage processes a stream of micro-batches, typically chosen as the largest units that fit in device memory.
For simplicity, we describe the computation from the perspective of a generic stage $f$ with parameters $\vw$ and a single micro-batch, represented by its input and output features $\vs$ and $\vs_+$, omitting stage and micro-batch indices.
We also denote the function that takes the input feature $\vs$ and evaluate the loss value (\ie, the composition of downstream stages and the loss function) as simply $L(\vs_+)\in\R$.
At a high level, each stage repeatedly performs three operations for each micro-batch: it computes the forward signal $\vs_+ = f(\vw, \vs)$ and sends it downstream; it receives a downstream error signal $\ve_+$ and computes the backward signal $\ve$ together with the local parameter gradient $\vg$; and, in between, it waits for downstream stages to complete the dependent forward or backward computations.

\paragraph{A close look at asynchronous pipelining}
Writing the asynchronous pipeline computation for a generic stage and a single micro-batch reveals its characteristic structure:
\begin{center}
\begin{minipage}[t]{0.6\linewidth}
    \vspace{-1em}
    \centering
    \begin{align}
        & \vs_+ = f(\vw, \vs)
        && {\footnotesize\color{gray}\triangleright~\textit{Forward signal}}  \label{eq:async_fwd}\\
        & \hspace{-0.5em}
           {\color{gray}(\vw \to \textcolor{paramred}{\vw^+}, \text{ wait for } \partial_{\vs_+} L)}
        && {\footnotesize\color{gray}\triangleright~\textit{Asynchronous updates}} \label{eq:async_wait} \\
        & \ve_+ = \partial_{\vs_+} L(\vs_+)
        && {\footnotesize\color{gray}\triangleright~\textit{Returned error signal}} \label{eq:async_err} \\
        & \ve = \mathbf{\partial}_{\vs} f(\textcolor{paramred}{\vw^+}, \vs)^*[\ve_+]
        && {\footnotesize\color{gray}\triangleright~\textit{Backprop error signal}} \label{eq:async_bwd} \\
        & \vg = \mathbf{\partial}_{\vw} f(\textcolor{paramred}{\vw^+}, \vs)^*[\ve_+]
        && {\footnotesize\color{gray}\triangleright~\textit{Compute gradient}}  \label{eq:async_grad} \\
        & \Delta = - U(\vg)
        && {\footnotesize\color{gray}\triangleright~\textit{Compute update}}\label{eq:async_upd},
    \end{align}
\end{minipage}
\hfill
\begin{minipage}[t]{0.33\linewidth}
    \centering
    \vspace{0pt}
    \includegraphics[width=0.75\linewidth,trim={1.3em 0 0 0},clip]{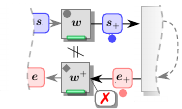}
    \captionof{figure}{Parameter-inconsistent gradient computation.}
    \label{fig:pipelined-gradient-lifecycle}
\end{minipage}
\end{center}
where $\mathbf{\partial}_{\vs} f(\vw^+,\vs)^*[\ve_+]$ and $\mathbf{\partial}_{\vw} f(\vw^+,\vs)^*[\ve_+]$ denote the reverse-mode chain-rule operations.

One property exposed by this formulation is how inconsistency becomes more severe for earlier pipeline stages.
The wait in (\ref{eq:async_wait}) accumulates upstream: before an upstream stage receives its returned error signal, the neighboring downstream stage must itself complete its waiting interval, together with the additional forward and backward computations in (\ref{eq:async_fwd}--\ref{eq:async_grad}).
As shown in \cref{fig:async-gantt}, this interval is long enough for the upstream stage to process other micro-batches and apply intervening updates, so the parameter state may drift from $\vw$ to $\vw^+$ before the original micro-batch returns for backpropagation in (\ref{eq:async_err}), illustrated in  \cref{fig:pipelined-gradient-lifecycle}.

\paragraph{Forward-equivalence condition}
Attending to the gradient computation in (\ref{eq:async_grad}), we identify the inconsistency between the returned error signal $\ve_+$ and the parameter state used in the local Jacobian $\partial_\vw f(\vw^+, \vs)$, which generally prevents them from forming a proper derivative chain.
This raises a natural question: \emph{can the same error signal $\ve_+$ remain valid across different parameter states $\vw^+$ that induce different local Jacobians?}
The key observation is that $\ve_+$ depends on the downstream feature $\vs_+$, not directly on the parameter tensor $\vw$ itself.
Therefore, if the updated parameter $\vw^+$ produces the same forward feature for the in-flight micro-batch, then the returned error signal can still be chained with the local Jacobian at $\vw^+$:
\begin{tcolorbox}[
    enhanced,
    colback=paramred!10,
    colframe=paramred!50,
    boxrule=1pt,
    arc=2pt,
    boxsep=2pt,
    left=5pt,
    right=5pt,
    top=5pt,
    bottom=5pt,
    before skip=6pt,
    after skip=6pt
]
\centering
$\displaystyle
\forall \vw^+:
f(\vw^+,\vs)=f(\vw,\vs)=\vs_+
\quad \Longrightarrow \quad
\vg =
\mathbf{\partial}_{\vw} f(\vw^+,\vs)^*[\ve_+]
\;\text{is a proper gradient at } \vw^+
$
\end{tcolorbox}
This forward-equivalence condition is the key idea behind our work: the full parameter tensor need not remain fixed, as long as the stage preserves the forward behavior seen by the in-flight micro-batch until its error signal returns.
Under this condition, the returned error signal and the changed local Jacobian still form a valid gradient at the updated parameter state.
The remaining question is therefore how to constrain asynchronous parameter updates so that they preserve these boundary signals while still allowing the parameters to evolve.

\section{Method}
\label{sec:method}

Building on this insight, we design \underline{F}o\underline{r}ward-\underline{E}quivalent \underline{S}ubspace-\underline{C}onstrained \underline{O}ptimization, or \ours, as the solution to the constrained optimization problem described below.
The resulting update rule exploits a natural non-interference structure of linear operators: updates that annihilate the in-flight activation subspace do not change the corresponding forwarded features.

\subsection{Derivation}
\label{sec:algorithm}

Consider a linear layer with bias absorbed into the weight matrix: $Y=W\bar X$, where $W\in\mathbb{R}^{d_{\mathrm{out}}\times(d_{\mathrm{in}}+1)}$.
For each such layer, we solve the constrained optimization problem
\begin{equation}
    \min_{\Delta'}
    \frac{1}{2}\|\Delta'-\Delta\|_F^2
    \quad
    \mathrm{s.t.}
    \quad
    (W+\Delta')\bar X = W\bar X,
\end{equation}
where $\bar X\in \mathbb{R}^{(d_{\mathrm{in}}+1)\times \tau n}$ denotes the concatenated input features of all $\tau$ in-flight micro-batches, each of size $n$, whose forward computations are to be protected from asynchronous parameter updates $\Delta\in\mathbb{R}^{d_{\mathrm{out}}\times(d_{\mathrm{in}}+1)}$.
Thus, the goal is to construct the feasible update $\Delta'$ closest in Frobenius norm $\|\,\cdot\,\|_F$ to the original update $\Delta$ such that the updated weight $W+\Delta'$ preserves the same forward pass as $W$ on the protected micro-batch.
If every asynchronous update that intersects the lifetime of this micro-batch is subjected to the same constraint, then the final updated weight $W^+$ remains forward-equivalent to the weight state under which the micro-batch entered protection.

Rewriting the preservation condition as
\begin{equation} \label{eq:non-interfere-upd}
    \Delta' \bar X = 0,
\end{equation}
shows that the constraint does not explicitly depend on the weight $W$; it only depends on the action of the update on the protected in-flight activations.

We solve the constrained problem through the KKT conditions.
Introducing a matrix-valued Lagrange multiplier
$\Lambda\in\mathbb R^{d_{\mathrm{out}}\times \tau n}$ with the same shape as $\Delta'\bar X$, the Lagrangian is
\[
    \mathcal L(\Delta',\Lambda)
    =
    \frac{1}{2}\|\Delta'-\Delta\|_F^2
    +
    \langle \Lambda,\Delta'\bar X\rangle_F,
\]
where $\langle A,B\rangle_F=\operatorname{tr}(A^\top B)$ denotes the Frobenius inner product.
Because the objective is convex quadratic and the constraint is affine, the KKT conditions are necessary and sufficient:
\begin{equation}
    \Delta'\bar X = 0,
    \qquad
    \nabla_{\Delta'}\mathcal L(\Delta',\Lambda)=0.
    \label{eq:kkt}
\end{equation}
The stationarity condition in \cref{eq:kkt} gives
\begin{equation}
    0
    =
    \Delta'-\Delta+\Lambda\bar X^\top,
    \qquad
    \Delta'=\Delta-\Lambda\bar X^\top.
    \label{eq:delta-prime-lambda}
\end{equation}
Substituting \cref{eq:delta-prime-lambda} into the feasibility condition gives $0 = (\Delta-\Lambda\bar X^\top)\bar X \notag  = \Delta\bar X-\Lambda\bar X^\top\bar X$, and therefore $\Lambda\bar X^\top\bar X=\Delta\bar X$.
Using the Moore--Penrose pseudoinverse\footnote{Although $\Lambda$ may not be unique when $\bar X^\top\bar X$ is singular, all valid choices yield the same product $\Lambda\bar X^\top$, and hence the same primal solution $\Delta'_\star$.}, we write $ \Lambda = \Delta\bar X(\bar X^\top\bar X)^\dagger$.
Finally, substituting this into \cref{eq:delta-prime-lambda} yields
\begin{equation}
    \tcboxmath[myeqbox]{
    \Delta'_\star
    =
    \Delta
    \left(
        I-\bar X(\bar X^\top\bar X)^\dagger\bar X^\top
    \right).
    }
    \label{eq:delta-star-projection}
\end{equation}

Here, the factor $\bar X(\bar X^\top\bar X)^\dagger\bar X^\top$ in \cref{eq:delta-star-projection} is the orthogonal projector onto $\operatorname{col}(\bar X)$.
Using the compact SVD $\bar X=U\Sigma V^\top$, we can write
$
    P_{\bar X}
    :=
    \bar X(\bar X^\top\bar X)^\dagger\bar X^\top
    =
    UU^\top.
$
Thus,
\begin{equation}
    \Delta'_\star
    =
    \Delta(I-P_{\bar X})
    =
    \Delta(I-UU^\top).
    \label{eq:fresco_projected_update_projector}
\end{equation}

\subsection{Relaxation}
\label{sec:relaxation}

For extremely deep pipelines, concurrency can reach a regime where each parameter update affects an increasing number of in-flight micro-batches, posing distinct challenges to both existing methods and \ours{}.
While existing methods suffer from increasingly stale gradients or unreliable predictions of future parameter states, \ours{} may find that protecting all affected micro-batches leaves no permissible direction for parameter movement, potentially annihilating the update altogether, \ie, $\Delta_\star'=0$.

Existing correction-based methods are not inherently equipped to handle this regime and often resort to auxiliary mechanisms such as step-size damping or momentum to maintain stability \citep{yang2021pipemare,ajanthan2025nesterov}.
Unlike these methods, \ours{} offers an explicit trade-off between gradient fidelity and effective parameter movement by relaxing the protection condition to preserve the most influential directions while allowing greater freedom for asynchronous parameter updates.

We do this by protecting only the spectrally dominant components of $\operatorname{col}(\bar X)$.
Specifically, we construct a rank-$k$ protection matrix $\bar X_k$ as the best low-rank approximation of $\bar X$ in the Frobenius norm: $\bar X_k = \arg\min_{\operatorname{rank}(A)\leq k} \|\bar X-A\|_F$.
Substituting this approximation into \ours{} yields the projection matrix $P_k=U_kU_k^\top$, where $U_k$ contains the top-$k$ left singular vectors of $\bar X$.
The resulting rank-limited update is
\begin{equation}
    \Delta'_k
    =
    \Delta(I-U_kU_k^\top).
    \label{eq:rank-k-projected-update}
\end{equation}
As the rank budget $k$ increases, the relaxation approaches exact preservation; when $k=\operatorname{rank}(\bar X)$, \cref{eq:rank-k-projected-update} recovers the exact projection in \cref{eq:fresco_projected_update_projector}.

Since weight matrices vary in dimensions $d_{\mathrm{out}}$ and $d_{\mathrm{in}}$, their protection capacities $\min(d_{\mathrm{out}},d_{\mathrm{in}}+1)$ also differ.
Using a fixed rank across layers may therefore lead to disproportionate protection or unnecessarily restrict parameter updates.
Instead, we allocate a fraction $\alpha\in[0,1]$ of each layer's capacity, capped by the maximum number $\tau n$ of in-flight feature directions:
\begin{equation}
    k=\min\left\{
    \tau n,\,
    \left\lceil\alpha\min(d_{\mathrm{out}},d_{\mathrm{in}}+1)\right\rceil
    \right\}.
\end{equation}
This provides a consistent relative protection budget across layers without allocating more directions than the in-flight features can span.
In practice, $U_k$ is assembled from per-micro-batch bases rather than computed directly from $\bar X$, as described in \cref{sec:algorithmic-description}.

\subsection{Algorithm}
\label{sec:algorithmic-description}

Here, we describe the algorithmic operation at a high level; the full procedure is given in \cref{alg:fresco-step}.
For each linear layer, \ours{} maintains a FIFO cache of low-rank activation bases for the in-flight micro-batches.
Since the protected rank $k$ is shared among the $\tau$ in-flight micro-batches, each micro-batch contributes $k_1=\lceil k/\tau\rceil$ directions, so that the cache holds $\tau k_1\geq k$ candidate directions.
When a new micro-batch enters the layer, \ours{} calls $\textsc{ExtractBasis}(\bar X,k_1)$ to compute a rank-$k_1$ basis of its bias-augmented input activations and enqueues the result.
We use an inexpensive approximation: a single step of block power iteration, orthogonalized by a single Newton--Schulz step (\cref{app:subspace-routines}).
When the corresponding backward error signal returns, that basis is dequeued, since the micro-batch no longer needs protection.

Before applying an optimizer update, \ours{} calls $\textsc{BuildProtectedBasis}(\mathcal Q,k)$, which concatenates the queued bases and keeps their dominant $k$ directions by block power iteration.
This produces a single orthonormal basis $Q\in\mathbb R^{(d_{\mathrm{in}}+1)\times k}$, the rank-$k$ protected subspace shared by the in-flight micro-batches.
The base optimizer first proposes the usual update $\Delta$; \ours{} then applies $\Delta'=\Delta(I-QQ^\top)$ instead, removing only the components that would change the protected forward features of the still in-flight micro-batches.
Because each micro-batch stores only $\lceil k/\tau\rceil$ directions, the cache holds fewer than $k+\tau$ basis vectors per layer regardless of the pipeline depth, and it stores only compact activation bases rather than delayed weight copies or predicted parameter buffers.
Thus, the main overhead is computation for subspace extraction, basis construction, and projection, rather than model-scale memory.

\begin{algorithm}[t]
\caption{\textbf{\ours} step}
\label{alg:fresco-step}
\begin{algorithmic}[1]
\Require Weight $W\in\mathbb R^{d_{\mathrm{out}}\times(d_{\mathrm{in}}+1)}$ (bias absorbed), input activation $X$, gradient $G$, protection budget $\alpha$, number $\tau$ of micro-batches in flight at this stage, mode $\in\{\textsc{Forward},\textsc{Backward}\}$
\Statex \textbf{State:} FIFO queue $\mathcal Q$ storing protected activation bases

\If{$\mathcal Q$ is uninitialized}
    \State Initialize $\mathcal Q\gets\emptyset$
\EndIf

\If{$\textit{mode}=\textsc{Forward}$}
    \State $\bar X \gets [X;\mathbf 1^\top]\in\mathbb R^{(d_{\mathrm{in}}+1)\times n}$
    \Comment{$n=$ batch $\times$ sequence}
    \State $k\gets \min\!\left(\tau n,\,\left\lceil\alpha \min(d_{\mathrm{out}},d_{\mathrm{in}}+1)\right\rceil\right)$
    \Comment{protected rank}
    \State $k_1\gets \lceil k/\tau\rceil$
    \Comment{rank per micro-batch}
    \State $V \gets \textsc{ExtractBasis}(\bar X,k_1)$
    \Comment{$V\in\mathbb R^{(d_{\mathrm{in}}+1)\times k_1}$}
    \State $\mathcal Q.\textsc{enqueue}(V)$
    \State \Return
\EndIf

\If{$\textit{mode}=\textsc{Backward}$}
    \State $\Delta\gets \textsc{OptimizerUpdate}(W,G)$
    \Comment{base optimizer proposes update}
    \State $\mathcal Q.\textsc{dequeue}()$
    \Comment{remove the completed micro-batch basis}
    \State $Q\gets \textsc{BuildProtectedBasis}(\mathcal Q,k)$
    \Comment{top-$k$ subspace of the remaining bases}
    \State $\Delta'\gets \Delta(I-QQ^\top)$
    \Comment{project away feature-changing directions}
    \State $W\gets W+\Delta'$
    \Comment{apply forward-equivalent update}
\EndIf
\end{algorithmic}
\end{algorithm}

\section{Theory}
\label{sec:theory}

We provide a preliminary convergence analysis of \ours{} under asynchronous pipeline execution.
Building on the perturbed forward--backward SGD framework of \citet{kong2026convergence}, we examine how the forward perturbations induced by concurrent parameter updates contribute to gradient errors and convergence.

For a linear stage, the residual forward perturbation of a projected update is
\begin{equation}
    R_{i,t}:=\|\Delta W'_{i,t}\bar S_{i,t}\|_F,
\end{equation}
where $\bar S_{i,t}$ contains the protected in-flight features.
For rank-$k$ projection, this residual is bounded by
\begin{equation}
    R_{i,t}\le\sigma_{i,t,k+1}\|\Delta W'_{i,t}\|_F,
\end{equation}
which vanishes when the complete activation subspace is protected.
Under bounded-delay execution, these residuals can be related to asynchronous gradient errors through the perturbation bounds of \citet{kong2026convergence}.

To examine the convergence implications, we consider a soft extension of \ours{},
\begin{equation}
    \Delta W'_{i,t}=\Delta W_{i,t}A_{i,t},
    \qquad
    A_{i,t}:=(I+\lambda_i U_{i,t,k}\Sigma_{i,t,k}^2U_{i,t,k}^\top)^{-1},
\end{equation}
which approaches hard projection as $\lambda_i\to\infty$.

\begin{theorem}[Convergence under soft projection]
\label{thm:fresco-convergence}
Under the assumptions in \cref{app:fresco-convergence-proof}, consider the update
$\vw_{t+1}=\vw_t-\eta\mathcal A_t\widetilde{\vg}_t$,
where $\mathcal A_t$ collects the soft projection operators and satisfies $aI\preceq\mathcal A_t\preceq I$ for some $a>0$.
For $\eta=c/\sqrt T$ with $0<c\le1/(4L)$,
\begin{equation}
\begin{aligned}
    \frac1T\sum_{t=0}^{T-1}
    \mathbb E\|\nabla\mathcal L(\vw_t)\|^2
    \le{}&
    \frac{2\Delta_{\mathcal L}+4Lc^2\sigma^2}{ac\sqrt T}
    +\frac{3c^2(K_{\mathrm f}s^2+K_{\mathrm b})}{aT},
\end{aligned}
\label{eq:fresco-convergence-bound}
\end{equation}
where $\Delta_{\mathcal L}:=\mathcal L(\vw_0)-\mathcal L_\star$, $s$ bounds the residual forward sensitivity, and $K_{\mathrm f},K_{\mathrm b}$ depend on the computational graph and pipeline delays.
\end{theorem}

This suggests that controlling forward perturbations is compatible with the standard $O(T^{-1/2})$ stochastic convergence rate under the stated conditions.
The guarantee is limited to the simplified soft-projection model and does not establish convergence acceleration or a full-gradient guarantee for hard \ours{}.
Details are provided in \cref{app:fresco-convergence-proof}.

\section{Evaluation}
\label{sec:experiments}

We compare \ours{} with a parameter-inconsistent baseline and five representative asynchronous pipeline-parallel strategies: parameter stashing~\citep{narayanan2019pipedream}, parameter recovery~\citep{yang2021pipemare}, and three parameter-speculation baselines, SpecTrain, PipeOptim, and LWP~\citep{chen2018efficient,guan2025pipeoptim,kosson2021pipelined}, which are described in detail in \cref{sec:related_async}.
Following prior study\cite{ajanthan2025nesterov}, we evaluate all methods on next-token prediction by pre-training a randomly initialized 95M-parameter NanoGPT decoder-only Transformer~\citep{Karpathy2022}.
The model is partitioned across $P=1$ to $32$ pipeline stages and trained on OpenWebText (OWT)~\citep{Gokaslan2019OpenWeb} with a micro-batch size of 8 and a sequence length of 512 for 20k iterations.
Additional experimental details are presented in \cref{app:exp-details}.
The code to reproduce the results of the paper is provided in \url{https://github.com/LOG-postech/fresco}.

\begin{figure*}[t]
    \centering

    \begin{subfigure}{0.23\linewidth}
        \centering
        \includegraphics[width=\linewidth]{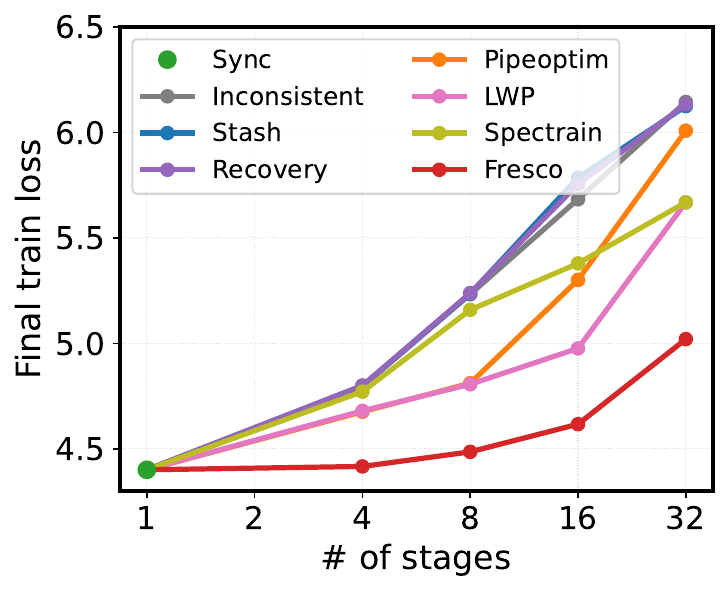}
        \caption{Depth degradation}
        \label{fig:main_combo_b}
    \end{subfigure}
    \hfill
    \begin{subfigure}{0.23\linewidth}
        \centering
        \includegraphics[width=\linewidth]{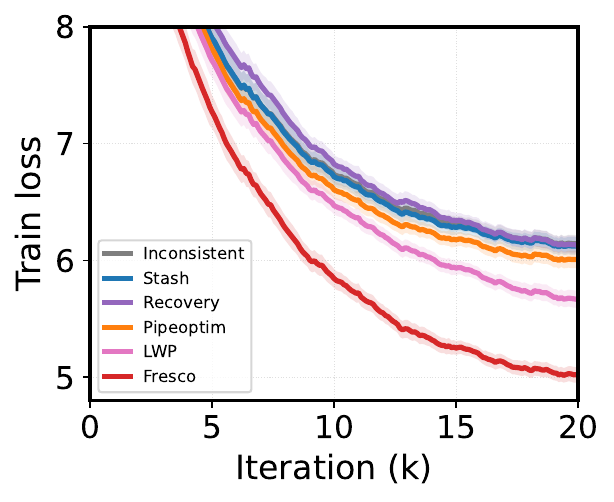}
        \caption{$P{=}32$ training curve}
        \label{fig:main_combo_a}
    \end{subfigure}
    \hfill
    \begin{subfigure}{0.23\linewidth}
        \centering
        \includegraphics[width=\linewidth]{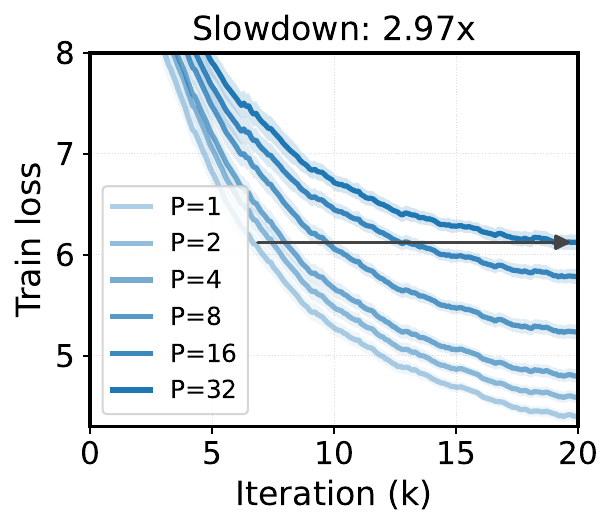}
        \caption{Parameter stashing}
        \label{fig:main_combo_d}
    \end{subfigure}
    \hfill
    \begin{subfigure}{0.23\linewidth}
        \centering
        \includegraphics[width=\linewidth]{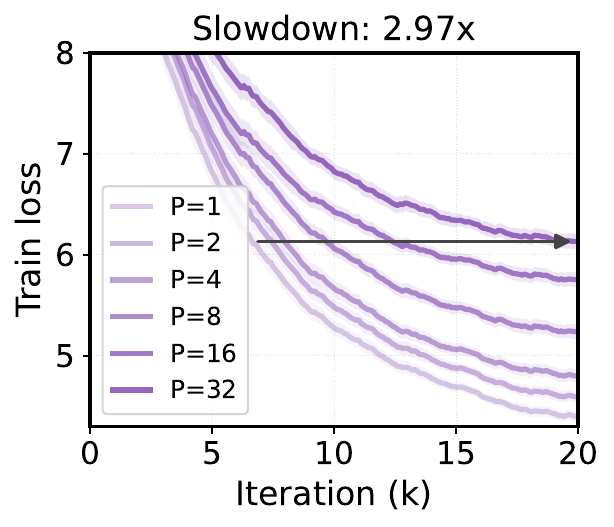}
        \caption{Parameter recovery}
        \label{fig:main_combo_e}
    \end{subfigure}

    \vspace{0.8em}

    \begin{subfigure}{0.23\linewidth}
        \centering
        \includegraphics[width=\linewidth]{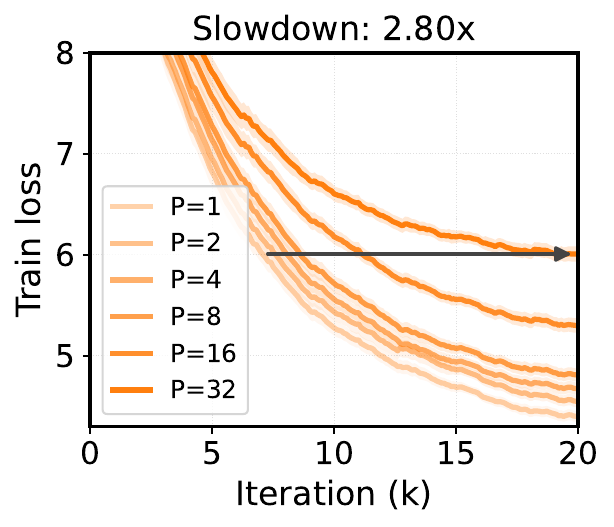}
        \caption{PipeOptim}
        \label{fig:main_combo_f}
    \end{subfigure}
    \hfill
    \begin{subfigure}{0.23\linewidth}
        \centering
        \includegraphics[width=\linewidth]{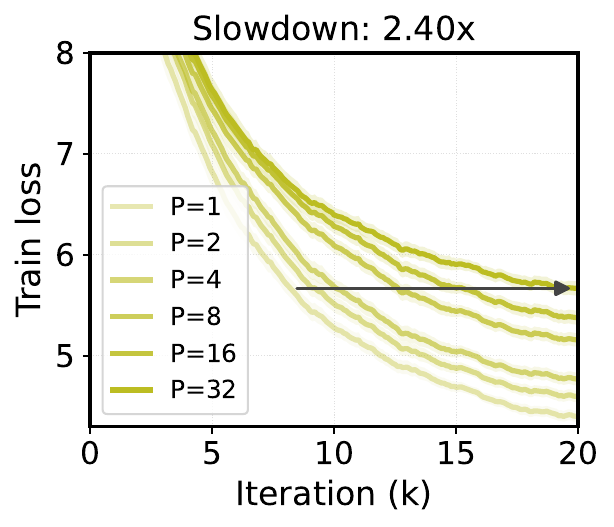}
        \caption{SpecTrain}
        \label{fig:main_combo_spectrain}
    \end{subfigure}
    \hfill
    \begin{subfigure}{0.23\linewidth}
        \centering
        \includegraphics[width=\linewidth]{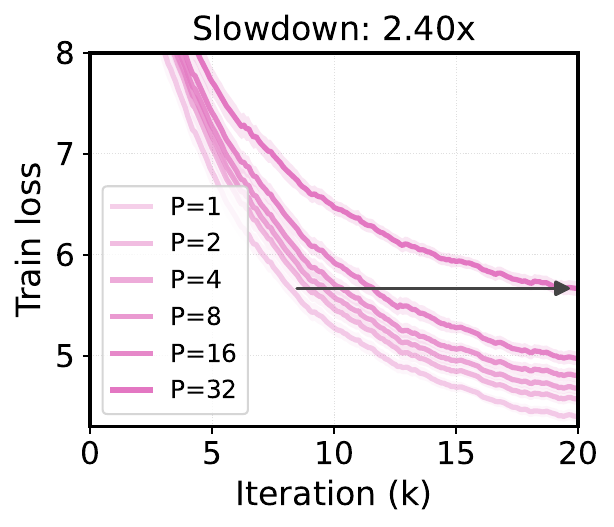}
        \caption{LWP}
        \label{fig:main_combo_g}
    \end{subfigure}
    \hfill
    \begin{subfigure}{0.23\linewidth}
        \centering
        \includegraphics[width=\linewidth]{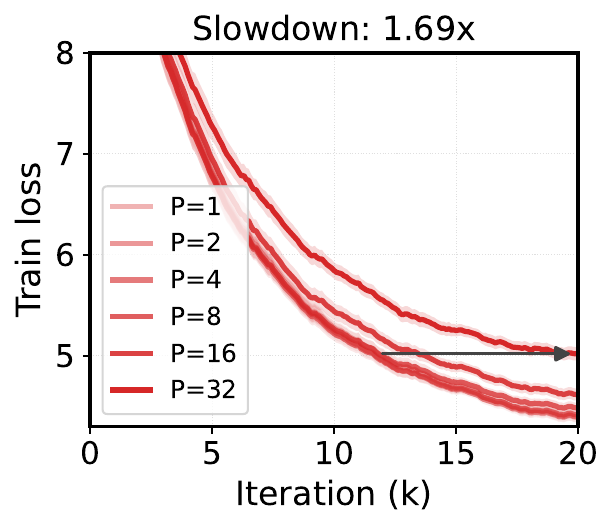}
        \caption{\ours{}}
        \label{fig:main_combo_h}
    \end{subfigure}

    \caption{
    Main results on 95M language-model pre-training.
    (a) \ours{} remains robust performance under deeper pipeline compared to other baselines, with (b) prominent optimization performance at $P=32$.
    (c--h) training curves across pipeline depths similarly demonstrates robustness of \ours{}.
    }
    \label{fig:main_result_combined}
\end{figure*}

\subsection{Robust optimization under deep pipelines}
We evaluate the robustness of \ours{} under increasingly severe inconsistency induced by deep asynchronous pipelining.
Specifically, we measure how well each method preserves optimization speed relative to the shallow-pipeline reference as the number of pipeline stages increases up to $P=32$.
\cref{fig:main_combo_b} reports the final training loss as the number of pipeline stages increases.
Most asynchronous baselines degrade sharply with larger $P$, indicating that their consistency mechanisms become increasingly insufficient as more micro-batches are simultaneously in flight.
By contrast, \ours{} maintains a substantially lower final loss under deep pipelines.
As shown in \cref{fig:main_combo_a}, at $P=32$, \ours{} converges substantially faster than the strongest baseline, LWP, reaching the same training loss with $43.2\%$ fewer iterations.

Additionally, the remaining panels of \cref{fig:main_result_combined} provide the full training curves across pipeline depths for each method.
We report slowdown as the relative number of iterations required to reach a fixed target loss compared with the shallow-pipeline reference.
Fully inconsistent asynchronous training, parameter stashing, parameter recovery, and PipeOptim all show slowdowns of approximately $2.80$--$2.97\times$, while SpecTrain and LWP still slow down by $2.40\times$ because they cannot perfectly predict the future trajectory.
In contrast, \ours{} reduces the slowdown to $1.69\times$.
Thus, even though \ours{} uses the current updated parameters rather than reconstructing or predicting a matched full parameter state, subspace-level preservation provides stronger robustness under deep asynchrony.

These results suggest that \ours{} establishes a more reliable strategy than prior baselines, supporting forward-equivalent update projection as effective in preserving optimization progress, avoiding both the stale-gradient limitation of parameter stashing and recovery and the instability of prediction-based methods under uncontrolled asynchronous updates to forward features that pending backward signals still depend on.

\begin{figure*}[t]
    \centering
    \begin{subfigure}{0.32\linewidth}
        \centering
        \includegraphics[width=\linewidth]{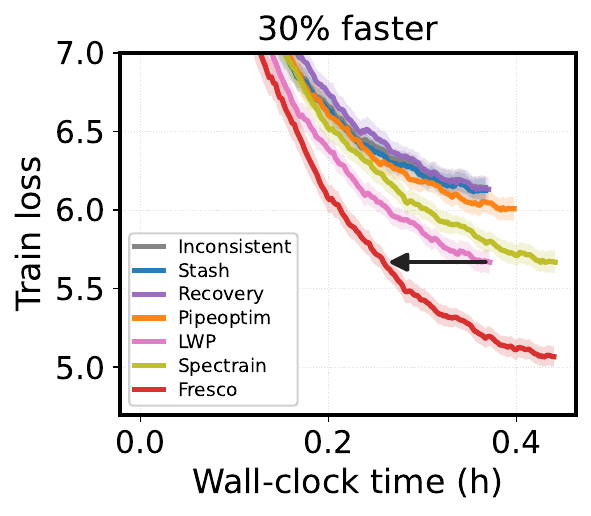}
        \caption{Loss vs. wall-clock time}
        \label{fig:hw_a}
    \end{subfigure}
    \hfill
    \begin{subfigure}{0.34\linewidth}
        \centering
        \includegraphics[width=\linewidth,trim={0 -0.9em 0 0},clip]{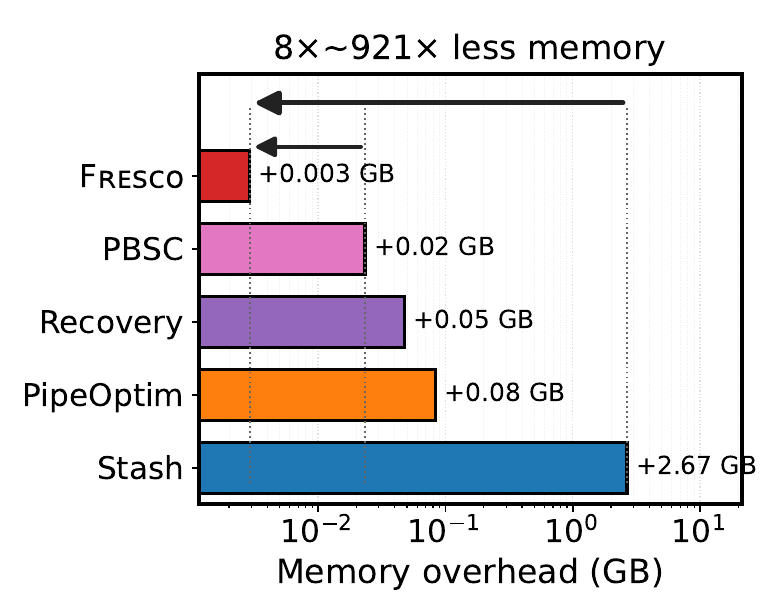}
        \caption{Memory overhead}
        \label{fig:hw_b}
    \end{subfigure}
    \hfill
    \begin{subfigure}{0.32\linewidth}
        \centering
        \includegraphics[width=\linewidth]{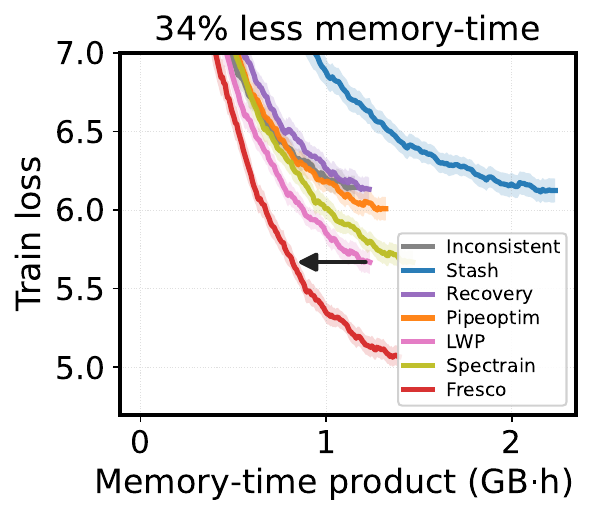}
        \caption{Loss vs. memory-time}
        \label{fig:hw_c}
    \end{subfigure}
    \caption{
    Compute efficiency of \ours{}.
    \ours{} improves both runtime and memory efficiency under deep pipelining: it reaches the target loss about (a) $30\%$ faster in wall-clock time, uses (b) from $8\times$ up to $921\times$ less memory overhead, and achieves a (c) $34\%$ lower memory-time product.
    These gains come from storing compact activation-subspace summaries rather than delayed full-parameter copies.
    }
    \vspace{-1em}
    \label{fig:hardware_combined}
\end{figure*}

\subsection{Compute efficiency}

We next evaluate the compute efficiency of \ours{} in terms of wall-clock time, memory overhead, and their combined memory-time product on $P=32$.
\cref{fig:hw_a} shows that \ours{} reaches the same target loss about $30\%$ faster than the strongest baseline in wall-clock time, despite the additional computation required for subspace estimation, basis merging, and update projection.
Also, the total memory overhead added by each asynchronous pipeline methods in \cref{fig:hw_b} show a complementary advantage.
This is due to all baselines rely on parameter-sized auxiliary buffers that is much larger than what \ours{} rely on, with parameter stashing additionally scaling with the number of pipeline stages by storing full parameter versions, reaching about $2.67$ GB in our setting.
In contrast, \ours{} stores only compact spectral summaries of in-flight activation subspaces, requiring about $3.0$ MB of additional memory, corresponding to a $921\times$ reduction over stashing.
This also translates into the end-to-end memory-time view in \cref{fig:hw_c}, where the memory-time product measures the cumulative memory footprint required to reach a target loss.
By this metric, \ours{} achieves the same loss with a $34\%$ lower memory-time product, showing that its gains do not merely shift cost from memory to additional training time.

\section{Discussion}
\label{sec:discussion}

\subsection{Protection capacity}
\label{sec:disc-capacity}

Any asynchronous pipeline-parallel method faces a finite consistency budget: as pipeline depth grows, more in-flight micro-batches must be made compatible with later parameter states, increasing either staleness, prediction error, memory, or update restriction.
Unlike prior asynchronous pipeline-parallel strategies, \ours{} provides explicit insight into how this capacity scales with pipeline depth: the protected rank should remain well below the layer input dimension, otherwise the feasible update space becomes too small.
As a rough estimate, if DeepSeek-V3, with its 61 Transformer blocks and smallest capacity being the expert intermediate width $2048$, is considered under its 4K pre-training sequences~\citep{liu2024deepseek}, this would imply that even one delayed batch-size-one micro-batch contains $4096$ token features, exceeding the $2048$-dimensional input of restrictive expert projections.
However, optimistically, recent practical measurements indicate that transformer activation spectra concentrate most of their mass in the top singular directions, fitting $\lambda_j \propto j^{-2\rho}$ with $2\rho \approx 1.2$--$1.3$~\citep{ferbach2026logarithmic}, under which the top $64$ directions in $7168$ dimensions capture about $72\%$--$80\%$ of the spectral mass.
Thus, while raw sequence length gives a pessimistic upper bound on the consistency burden, these empirical findings suggest that the effective burden may be much smaller in practice if spectral concentration generalizes, giving reason for optimism about scaling \ours{} and motivating the relaxed protection schemes proposed in \cref{sec:relaxation}.

\subsection{Trade-off between protection and progress}
\label{sec:disc-tradeoff}

Complete protection is likely not always beneficial, even when it is possible.
Indeed, the underlying nature of asynchronous pipelining is not merely that each update must protect an increasing number of in-flight micro-batches, but that this protection is inherently in tension with the update's own contribution to optimization progress.
Thus, practical schemes that expose dials for this balance, such as relaxed versions of \ours, may be important for identifying configurations that maximize performance within a fixed training time.
Another possible remedy is to improve the trade-off itself by strengthening the update or derivative oracle, as recent directions such as Nesterov-style updates or basis-rotation methods suggest that stronger optimization techniques may be naturally more robust to inconsistency, perhaps because their local approximation remains useful over a larger region \citep{dozat2016incorporating,ajanthan2025nesterov,jung2026mitigating}. %
Together, these observations suggest that future asynchronous pipeline training methods should not treat consistency preservation and optimization as separate components, but should instead co-design the projected update rule, the degree of protection, and the optimizer dynamics, especially in regimes where deeper pipelines make update interference increasingly severe.

\subsection{Extensions}
\label{sec:disc-extensions}

One potential extension concerns the update components discarded by \ours{}.
Although these components interfere with in-flight forward computations, they may still contain useful optimization information.
Rather than discarding them permanently, one could accumulate and reapply them once the corresponding protection constraints expire, resembling error-feedback mechanisms in efficient-state optimizers and distributed optimization~\citep{stich2019error,robert2024ldadam}.
However, these delayed updates may themselves become stale, making their effective reuse a nontrivial problem.

Another extension is to study how \ours{} interacts with hybrid parallelism in large-scale training, including data, tensor, sequence/context, and recent paradigms such as SWARM parallelism~\citep{shoeybi2019megatron,korthikanti2023reducing,zeng2025distributed,ryabinin2023swarm}.
Subspace Networks~\citep{ramasinghesubspace} are particularly relevant, as they argue that training can proceed with low-rank micro-batch forward and error signals, suggesting a naturally favorable setting for \ours{}.
A larger-scale stress test across model sizes, pipeline depths, and hybrid-parallel configurations is therefore an important next step, though beyond our available compute.

Finally, a more comprehensive convergence analysis would need to account for the coupled dynamics between parameter changes, the resulting asynchronous gradient errors, and the effects of different update-control mechanisms on subsequent optimization.
Characterizing these feedback interactions through a dynamical-systems perspective remains an interesting direction for future work.

\section{Conclusion}
\label{sec:conclusion}

We proposed a novel and principled \emph{forward-equivalence condition} that directly controls the asynchronous updates, leading us to designing \ours{} based on constrained optimization that demonstrates consistently strong performance and efficiency under deep pipelines compared to existing representative baselines.
We believe this shift from parameter-level consistency provides a promising foundation for future high-utilization training systems for stable and efficient training of larger models and more resource-limited distributed environments.

\section*{Acknowledgment}
This work was partly supported by the Institute of Information \& communications Technology Planning \& Evaluation (IITP) grant funded by the Korean government (MSIT) (RS-2019-II191906, Artificial Intelligence Graduate School Program (POSTECH), RS-2026-25507427, Development of Efficient Architectures and Training Techniques for High-Performance Lightweight AI Models), the National Research Foundation of Korea(NRF) grant funded by the Korea government(MSIT) (Grant No. RS-2026-25612350), Project Title: Development of an End-to-End Design and Validation Framework for Mathematically Principled Next-Generation AI Architectures and Learning Algorithms), the Korea Basic Science Institute(National research Facilities and Equipment Center) grant funded by the Korea government(MSIT) (RS-2026-25500419).

\bibliography{references}
\bibliographystyle{plainnat}

\newpage
\appendix
\onecolumn

\section{Related works}

\subsection{Asynchronous pipeline parallelism methods}
\label{sec:related_async}

\paragraph{Retrospective consistency and staleness}
One dominant strategy is to compute gradients using the forward-time parameters to ensure consistency between the forward and backward passes \citep{narayanan2019pipedream, narayanan2021pipedream2bw, yang2022wpipe, yang2021pipemare}.
PipeDream \citep{narayanan2019pipedream} introduced \emph{parameter stashing}, which keeps a snapshot of the forward-time parameters for each micro-batch and uses it during the backward pass, analogous to process forking.
However, this approach incurs \emph{significant memory overhead}; in the worst case, the first pipeline stage must store parameters equivalent to the entire model.
PipeMare \citep{yang2021pipemare} addresses this limitation by maintaining an exponential moving average of gradients and applying a first-order Taylor approximation of the optimization trajectory to recover the parameters.
Although these approaches produce gradients consistent with the forward-time parameters, they remain stale relative to the latest parameters.

\paragraph{Prospective consistency and forecast error}
Another line of work predicts the parameters to be used at backward time instead of storing forward-time copies, reducing memory overhead at the cost of prediction error. \citep{chen2018efficient, guan2019xpipe}
SpecTrain \citep{chen2018efficient} performs lookahead-style parameter prediction using momentum-smoothed gradients to forecast future parameters at forward time.
XPipe \citep{guan2019xpipe} refines this by incorporating optimizer-aware update directions (e.g., Adam moments) to improve prediction accuracy and robustness.
DC-ASGD \citep{zheng2017asynchronous} instead compensates for delay at the gradient level by applying a first-order Taylor correction, approximating gradients at the current parameters.
Similarily to Pipemare, reliance on local approximations to extrapolate non-local parameter trajectories would depend on sufficiently smooth optimization trajectories (often induced by momentum \citep{ajanthan2025nesterov}) and may degrade under noisy or rapidly changing gradients.

\paragraph{Restricting optimization progress}
Another line of work regularizes updates to limit the deviation between forward- and backward-time parameters, reducing inconsistency without explicit prediction or stashing.
AvgPipe \citep{chen2023avgpipe} applies elastic averaging to keep local parameters close to a shared reference, effectively constraining drift across pipeline stages.
PipeMare \citep{yang2021pipemare} incorporates delay-aware learning rate rescheduling to dampen updates under larger staleness.
Momentum correction \citep{ajanthan2025nesterov} can be viewed as implicitly smoothing the optimization trajectory, reducing abrupt deviations.
However, these methods restrict update dynamics globally, which may limit optimization efficiency and do not directly resolve parameter inconsistency.

\subsection{Connection to asynchronous and semantic concurrency control}
\label{sec:related_semantic_locking}

Our work is conceptually connected to asynchronous numerical computation, which studies alternatives to conventional iterative methods that proceed without global synchronization under delayed or partially inconsistent updates~\citep{chazan1969chaotic,baudet1978asynchronous,frommer2000asynchronous,feyzmahdavian2023asynchronous}.
It is also distantly related to semantic concurrency control, where conflicts are defined by the abstract effects of operations rather than by low-level read--write overlap~\citep{weihl1988commutativity,weihl1989local,badrinath1992semantics,resende1994semantic}.
We draw only on the shared abstraction principle: synchronization need not always be imposed on the full memory representation when the semantics of the computation identify a smaller conflicting component.
Our method can be seen as a domain-specific instantiation of this principle in pipeline-parallel continuous optimization: consistency is enforced on the semantically relevant effect of an update, rather than on its full parameter representation.

\subsection{Connection to gradient projection methods in continual and multi-task learning}
Projecting parameter updates onto the orthogonal complement of a protected feature subspace recurs across several areas of literature.
In \emph{continual learning}, OWM~\citep{zeng2018continual}, OGD~\citep{farajtabar2020orthogonal}, GPM~\citep{saha2021gradient}, and Adam-NSCL~\citep{wang2021training} constrain new-task updates to the null space of features from previous tasks, with later work relaxing this constraint or extending it to transformers and state-space models~\citep{lin2022trgp,lu2024visual,cheng2024mambacl}.
The same construction appears in \emph{knowledge editing}~\citep{fang2025alphaedit}, \emph{multi-task gradient surgery}~\citep{yu2020gradient,liu2021cagrad}, and \emph{machine unlearning}~\citep{chen2024unsc,hoang2024learn}.
The closest precedent is OrthoFL~\citep{zhang2025orthogonal}, which projects asynchronous federated-learning drift onto a subspace orthogonal to stale client updates.
In all of the above, the protected set is essentially stationary---fixed before the update---so the projector can be precomputed once and reused.
In our setting, the protected set is the in-flight micro-batches, which turns over every step; the projector must be rebuilt on the critical path under the per-stage budget of pipeline parallelism.%

\section{Proof of \cref{thm:fresco-convergence}}
\label{app:fresco-convergence-proof}

We extend the perturbed forward--backward SGD analysis of \citet{kong2026convergence} to a simplified model of \ours{}.
We first bound the perturbations induced by asynchronous parameter changes and then derive the convergence rate under soft projection.

\subsection{Model and Assumptions}

Consider a pipeline of $P$ stages with linear parameterized operators $f_i(\vw,\vs)=W_i\vs$.
For an in-flight micro-batch, let $W^{\mathrm f}_{i,t}$, $W^{\mathrm b}_{i,t}$, and $W^{\mathrm c}_{i,t}$ denote the parameters used during forward propagation, backward propagation, and gradient application, respectively.
The corresponding forward and backward perturbations are
\begin{equation}
\begin{aligned}
    \delta^{\mathrm f}_{i,t}
    &:=(W^{\mathrm f}_{i,t}-W^{\mathrm c}_{i,t})\vs_{i,t},\\
    \delta^{\mathrm b}_{i,t}
    &:=(W^{\mathrm b}_{i,t}-W^{\mathrm c}_{i,t})^\top\ve_{+,i,t}.
\end{aligned}
\label{eq:app-async-perturbations}
\end{equation}
These quantities measure the local computational discrepancies relative to the current parameters, holding the input feature and incoming error signal fixed.

At update event $u$, let $\bar S_{i,u}$ collect the protected in-flight features, with compact SVD
$\bar S_{i,u}=U_{i,u}\Sigma_{i,u}Q_{i,u}^\top$.
For analysis, we consider the soft-projection update
\begin{equation}
    \Delta W'_{i,u}=\Delta W_{i,u}A_{i,u},
    \qquad
    A_{i,u}:=(I+\lambda_{i,u}U_{i,u,k}\Sigma_{i,u,k}^2U_{i,u,k}^\top)^{-1}.
    \label{eq:app-soft-update}
\end{equation}
Here $\Delta W_{i,u}=-\eta G_{i,u}$ is the proposed update, with $G_{i,u}$ denoting the matrix representation of the asynchronously computed stage gradient.

The following assumptions are used.

\begin{assumption}[Smoothness]
\label{ass:fresco-smoothness}
The objective $\mathcal L$ is bounded below by $\mathcal L_\star$ and has an $L$-Lipschitz gradient:
\begin{equation}
    \|\nabla\mathcal L(\vw)-\nabla\mathcal L(\vw')\|
    \le L\|\vw-\vw'\|.
\end{equation}
\end{assumption}

\begin{assumption}[Stochastic gradients]
\label{ass:fresco-stochasticity}
Let $\mathcal F_t$ denote the information available before sampling the ideal stochastic gradient, including the current parameters and update operator $\mathcal A_t$.
The unperturbed gradient $\vg_t$ satisfies
\begin{equation}
    \mathbb E[\vg_t\mid\mathcal F_t]
    =\nabla\mathcal L(\vw_t),
\end{equation}
and
\begin{equation}
    \mathbb E[\|\vg_t-\nabla\mathcal L(\vw_t)\|^2
    \mid\mathcal F_t]\le\sigma^2.
\end{equation}
\end{assumption}

\begin{assumption}[Operator smoothness]
\label{ass:fresco-operators}
For each operator $f_i$, let $D_1f_i$ and $D_2f_i$ denote its input and parameter Jacobians.
There exist finite constants $C_1,C_2,L_f,L_1,L_2>0$ such that, for all computational states $z=(\vs,\vw_i)$ and $z'=(\vs',\vw_i')$ under consideration,
\begin{equation}
    \|D_jf_i(z)\|_{\mathrm{op}}\le C_1,
    \qquad j\in\{1,2\},
\end{equation}
\begin{equation}
    \|D_{21}^2f_i(z)\|_{\mathrm{op}}\le C_2,
\end{equation}
\begin{equation}
    \|f_i(\vw_i,\vs)-f_i(\vw_i,\vs')\|
    \le L_f\|\vs-\vs'\|,
\end{equation}
and
\begin{equation}
\begin{aligned}
    \|Df_i(z)-Df_i(z')\|_F
    &\le L_1\|z-z'\|,\\
    \|D^2f_i(z)-D^2f_i(z')\|_F
    &\le L_2\|z-z'\|.
\end{aligned}
\end{equation}
These conditions apply to the differentiable computational graph, including the terminal loss operator.
\end{assumption}

\begin{assumption}[Bounded computational signals]
\label{ass:fresco-signals}
There exist finite constants $G_i,E_i,M_i>0$ such that, almost surely,
\begin{equation}
    \|G_{i,t}\|_F\le G_i,\qquad
    \|\ve_{+,i,t}\|\le E_i,\qquad
    \|\bar S_{i,t}\|_2\le M_i.
\end{equation}
\end{assumption}

Assumptions \ref{ass:fresco-smoothness}--\ref{ass:fresco-operators} follow the framework of \citet{kong2026convergence}.
The stochastic-gradient condition is strengthened to condition on the update operator, which is required for the projected descent argument and is not automatic when the projection depends on the gradient sample.
Assumption \ref{ass:fresco-signals} supplies uniform bounds on the quantities involved in asynchronous updates.

We consider a fixed pipeline schedule for which the number of intervening stage updates is bounded.
Let $\mathcal H_{i,t}$ and $\mathcal J_{i,t}$ denote the updates separating the forward and backward parameter versions from the current version.
The execution model determines finite constants $\tau_i^{\mathrm f},\tau_i^{\mathrm b}$ satisfying
\begin{equation}
    |\mathcal H_{i,t}|\le\tau_i^{\mathrm f},
    \qquad
    |\mathcal J_{i,t}|\le\tau_i^{\mathrm b}.
    \label{eq:app-delay}
\end{equation}

For the convergence analysis, we use a logical-update model that collects the stage-wise soft transformations into a block-diagonal operator $\mathcal A_t$:
\begin{equation}
    \vw_{t+1}
    =\vw_t-\eta\mathcal A_t\widetilde{\vg}_t.
    \label{eq:app-logical-update}
\end{equation}
This model retains the gradient perturbations induced by the pipeline while abstracting away the exact ordering of physical stage updates.

For $0\le\lambda_{i,t}\le\Lambda<\infty$, the eigenvalues of $A_{i,t}$ lie in
$[(1+\Lambda M^2)^{-1},1]$, where $M:=\max_iM_i$.
Hence
\begin{equation}
    aI\preceq\mathcal A_t\preceq I,
    \qquad
    a:=\frac{1}{1+\Lambda M^2}>0.
    \label{eq:app-operator-bounds}
\end{equation}
This bound follows from the soft-projection definition.

\subsection{Perturbation Bound and Convergence}

We first characterize the residual forward perturbation.
By the SVD of $\bar S_{i,u}$, \cref{eq:app-soft-update} gives
\begin{equation}
\begin{aligned}
    R_{i,u}^2
    &:=\|\Delta W'_{i,u}\bar S_{i,u}\|_F^2\\
    &=\sum_{j=1}^{k}
    \frac{\sigma_{i,u,j}^2\|\Delta W_{i,u}u_{i,u,j}\|^2}
    {(1+\lambda_{i,u}\sigma_{i,u,j}^2)^2}\\
    &\quad+\sum_{j=k+1}^{r_{i,u}}
    \sigma_{i,u,j}^2\|\Delta W_{i,u}u_{i,u,j}\|^2.
\end{aligned}
\label{eq:app-spectral-residual}
\end{equation}
Define
\begin{equation}
    s_{i,u}:=\max\left\{
    \max_{j\le k}
    \frac{\sigma_{i,u,j}}{1+\lambda_{i,u}\sigma_{i,u,j}^2},
    \ \sigma_{i,u,k+1}
    \right\},
    \label{eq:app-forward-sensitivity}
\end{equation}
with $\sigma_{i,u,r_{i,u}+1}=0$ and empty maxima equal to zero.
Then
\begin{equation}
    R_{i,u}\le s_{i,u}\|\Delta W_{i,u}\|_F
    \le\eta G_i s_{i,u}.
    \label{eq:app-residual-bound}
\end{equation}
Let $s:=\sup_{i,u}s_{i,u}\le M$.
For hard rank-$k$ projection, obtained as $\lambda\to\infty$, the first sum in \cref{eq:app-spectral-residual} vanishes.
In particular, protecting the complete activation subspace gives $R_{i,u}=0$.

\begin{lemma}[Asynchronous gradient error]
\label{lemma:fresco-error}
Under \cref{ass:fresco-operators,ass:fresco-signals} and the bounded-delay execution model, the gradient discrepancy
$\mathbf r_t:=\widetilde{\vg}_t-\vg_t$ satisfies
\begin{equation}
    \mathbb E\|\mathbf r_t\|^2
    \le\eta^2(K_{\mathrm f}s^2+K_{\mathrm b}),
    \label{eq:app-gradient-error}
\end{equation}
where $K_{\mathrm f}$ and $K_{\mathrm b}$ depend on the computational graph, update delays, and bounds on computational signals.
\end{lemma}

\begin{proof}
The difference between the forward and current parameter states is the sum of intervening updates.
Since each such update protects the feature of the affected in-flight micro-batch, \cref{eq:app-residual-bound} gives
\begin{equation}
\begin{aligned}
    \|\delta^{\mathrm f}_{i,t}\|
    &\le\sum_{u\in\mathcal H_{i,t}}R_{i,u}\\
    &\le\eta\tau_i^{\mathrm f}G_i s.
\end{aligned}
\label{eq:app-forward-error}
\end{equation}
Similarly, since $\|A_{i,u}\|_2\le1$,
\begin{equation}
\begin{aligned}
    \|\delta^{\mathrm b}_{i,t}\|
    &\le\|\ve_{+,i,t}\|
    \sum_{u\in\mathcal J_{i,t}}\|\Delta W'_{i,u}\|_2\\
    &\le\eta\tau_i^{\mathrm b}G_iE_i.
\end{aligned}
\label{eq:app-backward-error}
\end{equation}

By the perturbation-propagation result of \citet[Theorem~1]{kong2026convergence}, there exist nonnegative coefficients $C_i^{\mathrm f},C_i^{\mathrm b}$ such that
\begin{equation}
\begin{aligned}
    \mathbb E\|\mathbf r_t\|^2
    \le{}&
    \sum_i C_i^{\mathrm f}
    \mathbb E\|\delta^{\mathrm f}_{i,t}\|^2\\
    &+\sum_i C_i^{\mathrm b}
    \mathbb E\|\delta^{\mathrm b}_{i,t}\|^2.
\end{aligned}
\label{eq:app-error-propagation}
\end{equation}
Substituting \cref{eq:app-forward-error,eq:app-backward-error} proves the result with
\begin{equation}
\begin{aligned}
    K_{\mathrm f}
    &:=\sum_i C_i^{\mathrm f}(\tau_i^{\mathrm f}G_i)^2,\\
    K_{\mathrm b}
    &:=\sum_i C_i^{\mathrm b}(\tau_i^{\mathrm b}G_iE_i)^2.
\end{aligned}
\end{equation}
\end{proof}

The lemma bounds the squared gradient error by $O(\eta^2)$, with its forward component depending on the residual spectral sensitivity $s$.

\begin{proof}[Proof of \cref{thm:fresco-convergence}]
Write
\begin{equation}
    \widetilde{\vg}_t
    =\nabla\mathcal L(\vw_t)+\boldsymbol{\xi}_t+\mathbf r_t,
\end{equation}
where $\boldsymbol{\xi}_t:=\vg_t-\nabla\mathcal L(\vw_t)$ is the unperturbed stochastic-gradient noise.

By $L$-smoothness and \cref{eq:app-logical-update},
\begin{equation}
\begin{aligned}
    \mathcal L(\vw_{t+1})
    \le{}&\mathcal L(\vw_t)
    -\eta\langle\nabla\mathcal L(\vw_t),
    \mathcal A_t\widetilde{\vg}_t\rangle\\
    &+\frac{L\eta^2}{2}
    \|\mathcal A_t\widetilde{\vg}_t\|^2.
\end{aligned}
\label{eq:app-descent}
\end{equation}

Let $\mathbf h_t:=\nabla\mathcal L(\vw_t)$.
By conditional unbiasedness,
\begin{equation}
    \mathbb E[
    \langle\mathbf h_t,\mathcal A_t\boldsymbol{\xi}_t\rangle
    \mid\mathcal F_t]=0.
\end{equation}
Furthermore, Young's inequality gives
\begin{equation}
    -\langle\mathbf h_t,\mathcal A_t\mathbf r_t\rangle
    \le\frac14\|\mathcal A_t^{1/2}\mathbf h_t\|^2
    +\|\mathcal A_t^{1/2}\mathbf r_t\|^2,
\end{equation}
and
\begin{equation}
\begin{aligned}
    \|\mathcal A_t\widetilde{\vg}_t\|^2
    \le{}&
    2\|\mathcal A_t\mathbf h_t\|^2
    +4\|\mathcal A_t\boldsymbol{\xi}_t\|^2\\
    &+4\|\mathcal A_t\mathbf r_t\|^2.
\end{aligned}
\end{equation}

Using $aI\preceq\mathcal A_t\preceq I$ and $\eta\le1/(4L)$, taking expectations in \cref{eq:app-descent} yields
\begin{equation}
\begin{aligned}
    \frac{a\eta}{2}
    \mathbb E\|\nabla\mathcal L(\vw_t)\|^2
    \le{}&
    \mathbb E[\mathcal L(\vw_t)-\mathcal L(\vw_{t+1})]\\
    &+2L\eta^2\sigma^2
    +\frac{3\eta}{2}\mathbb E\|\mathbf r_t\|^2.
\end{aligned}
\label{eq:app-one-step-descent}
\end{equation}

Summing over $t=0,\ldots,T-1$, using
$\mathcal L(\vw_T)\ge\mathcal L_\star$, and applying \cref{lemma:fresco-error}, we obtain
\begin{equation}
\begin{aligned}
    \frac1T\sum_{t=0}^{T-1}
    \mathbb E\|\nabla\mathcal L(\vw_t)\|^2
    \le{}&
    \frac{2\Delta_{\mathcal L}}{a\eta T}
    +\frac{4L\eta\sigma^2}{a}\\
    &+\frac{3\eta^2(K_{\mathrm f}s^2+K_{\mathrm b})}{a},
\end{aligned}
\label{eq:app-convergence-intermediate}
\end{equation}
where $\Delta_{\mathcal L}:=\mathcal L(\vw_0)-\mathcal L_\star$.

Choosing $\eta=c/\sqrt T$ proves \cref{eq:fresco-convergence-bound}.
\end{proof}

The bound preserves the standard $O(T^{-1/2})$ nonconvex stochastic convergence rate, with an additional $O(T^{-1})$ term depending on the residual forward sensitivity and backward perturbations.
It does not imply that stronger projection necessarily improves convergence, since the minimum retained update gain $a$ also decreases with the protection strength.
The guarantee concerns the simplified soft-projection model; convergence of the implemented hard projection under the full asynchronous execution dynamics remains outside its scope.

\section{Experimental Details}
\label{app:exp-details}

\subsection{Model and Training Configuration}

Our experimental setup generally follows \citet{ajanthan2025nesterov}.
We use a nanoGPT-style decoder-only Transformer~\citep{Karpathy2022} with $L=32$ Transformer blocks, embedding dimension $d=384$, $a=6$ attention heads, and context length $s=512$, comprising approximately $95$M parameters.
The model uses learnable positional embeddings and an untied language modeling head.
At pipeline depth $P$, each stage contains $L/P$ Transformer blocks, with the embedding layer assigned to the first stage and the LM head to the last.

We train on OpenWebText for $20{,}000$ optimizer steps with micro-batch size $b=8$ and \texttt{FP32} precision.
We use NAdamW with weight decay $0.01$, gradient clipping at $1.0$, and a $3{,}000$-step linear warmup followed by cosine decay to $0.1$ times the peak learning rate.
We disable activation recomputation for all methods.
The common configuration is summarized in \cref{tab:common_hparams_95m_20k}.

\begin{table}[t]
\caption{\textbf{Common experimental configuration.}
Hyperparameters marked as swept are selected separately for each method and pipeline depth.}
\label{tab:common_hparams_95m_20k}
\centering
\small
\begin{tabular}{@{}ll@{}}
\toprule
\textbf{Configuration} & \textbf{Value} \\
\midrule
\rowcolor{gray!10}
\multicolumn{2}{@{}l}{\textbf{Model}} \\
Architecture & nanoGPT decoder-only Transformer \\
Embedding dimension $d$ & $384$ \\
Attention heads $a$ & $6$ \\
Transformer blocks $L$ & $32$ \\
Context length $s$ & $512$ \\
Parameters & $\sim 95$M \\
Positional embeddings & Learnable \\
LM head & Untied \\
\midrule
\rowcolor{gray!10}
\multicolumn{2}{@{}l}{\textbf{Training}} \\
Dataset & OpenWebText \\
Micro-batch size $b$ & $8$ \\
Precision & \texttt{FP32} \\
Optimizer & NAdamW \\
$\beta_1$ (swept) & $\{0.9,\,0.99\}$ \\
$\beta_2$, $\epsilon$ & $0.999$, $10^{-8}$ \\
Weight decay & $0.01$ \\
Gradient clipping & $1.0$ \\
Peak LR (swept) & $\{10^{-4},\,3\times10^{-4},\,10^{-3}\}$ \\
LR warmup & $3{,}000$ steps, linear \\
LR decay & Cosine to $0.1\times$ peak LR \\
Training horizon & $20{,}000$ steps \\
Random seed & $0$ \\
\midrule
\rowcolor{gray!10}
\multicolumn{2}{@{}l}{\textbf{Distributed Setup}} \\
Pipeline depths $P$ & $\{1,\,2,\,4,\,8,\,16,\,32\}$ \\
GPUs per run & $8$ \\
Training GPUs & RTX 3090 / RTX A6000 \\
Timing GPUs & $8\times$ RTX 3090 (24 GB) \\
Backend & \texttt{gloo} \\
Activation recomputation & Disabled \\
\bottomrule
\end{tabular}
\end{table}

\subsection{Hyperparameter Selection}

\paragraph{Learning rate and momentum.}
For each method and pipeline depth
$P\in\{1,2,4,8,16,32\}$, we sweep the peak learning rate and first-moment coefficient over
\begin{equation}
    \eta_{\max}\in\{10^{-4},\,3\times10^{-4},\,10^{-3}\},
    \qquad
    \beta_1\in\{0.9,\,0.99\}.
\end{equation}
Each configuration is evaluated with a single seed ($0$).
For the main results, we select the configuration with the lowest final training loss, measured by an exponential moving average (EMA) with decay factor $0.97$, evaluated at step $19{,}900$.

Ablations over the learning rate, $\beta_1$, learning-rate damping, and the protection budget are reported in \cref{app:ablation}.

\paragraph{Protection budget.}
We select the protection budget $\alpha$ at the deepest pipeline depth, $P=32$, over the full $20{,}000$-step training horizon.
Specifically, we sweep
\begin{equation}
    k\in\{12,\,24,\,48,\,64,\,96,\,154,\,269,\,384\},
    \qquad
    \alpha=\frac{k}{384}.
    \label{eq:app-alpha-search}
\end{equation}
For each linear layer, the protected rank is $\lceil\alpha\min(d_{\mathrm{out}},d_{\mathrm{in}})\rceil$, with bias augmentation adding an input coordinate, so $k=\alpha\cdot384$ for the $384$-dimensional layers ($k+1$ for the $385$-dimensional ones).
Each budget is evaluated using its best learning rate and $\beta_1$.
The results are shown in \cref{fig:alpha_p32}.

We select $\alpha=96/384=0.25$ based on the lowest evaluation loss.
Smaller budgets provide insufficient protection against forward interference, whereas larger budgets increasingly restrict the available update directions.
Full-rank protection can reach a lower training loss ($4.98$ at a peak LR of $10^{-3}$) but then generalizes poorly (evaluation loss $6.19$), which is why we select budgets by evaluation loss.
We use $\alpha=0.25$ for the pipeline-depth sweep.
For wall-clock comparisons, we instead use $\alpha=48/384=0.125$, which provides the best loss at a fixed wall-clock budget.

For Recovery, we similarly sweep the discrepancy-correction coefficient over
\begin{equation}
    D\in\{0.01,\,0.02,\,0.05,\,0.1\},
\end{equation}
and use $D=0.02$ throughout.

\subsection{Subspace Extraction and Projection}
\label{app:subspace-routines}

\ours{} uses bias-augmented projection: each input activation matrix gains a constant row of ones, $\bar X=[X;\mathbf 1^\top]\in\mathbb R^{d\times n}$ with $d=d_{\mathrm{in}}+1$, so that the projection also protects the bias term.
With the protected rank $k$ and $\tau$ in-flight micro-batches, each micro-batch contributes $k_1=\lceil k/\tau\rceil$ directions.

\paragraph{Subspace extraction.}
$\textsc{ExtractBasis}(\bar X,k_1)$ approximates the leading $k_1$ left singular vectors of $\bar X$ by block power iteration.
From an initial block $V_0\in\mathbb R^{d\times k_1}$, we iterate
\begin{equation}
    V_{j+1}
    =\operatorname{NS}_{T_{\mathrm{ns}}}(\bar X\bar X^\top V_j),
    \qquad j=0,\ldots,T_{\mathrm{bp}}-1,
\end{equation}
where $\operatorname{NS}_{T}$ normalizes its argument, $C\gets C/\|C\|_F$, and applies $T$ Newton--Schulz steps $C\gets\frac12 C(3I-C^\top C)$, and return $V=V_{T_{\mathrm{bp}}}$.
We use $T_{\mathrm{bp}}=1$ and $T_{\mathrm{ns}}=1$, so each micro-batch costs two multiplications by its activations and a few small matrix products.

\paragraph{Protected-basis construction.}
$\textsc{BuildProtectedBasis}(\mathcal Q,k)$ combines the bases $\mathcal Q=\{V^{(1)},\ldots,V^{(\tau)}\}$ of the in-flight micro-batches into a rank-$k$ orthonormal basis.
Their concatenation
\begin{equation}
    C=[V^{(1)},\ldots,V^{(\tau)}]
    \in\mathbb R^{d\times \tau k_1},
    \qquad \tau k_1\geq k,
\end{equation}
spans up to $\tau k_1$ directions.
When $C$ has at most $k$ columns, we take $Q=\operatorname{qr}(C)$; otherwise, we keep the dominant $k$ directions of $C$ by block power iteration from an initial block $Q_0\in\mathbb R^{d\times k}$,
\begin{equation}
    Q_{j+1}=\operatorname{qr}(CC^\top Q_j),
    \qquad j=0,\ldots,T_{\mathrm{c}}-1,
\end{equation}
with $T_{\mathrm{c}}=2$, where $\operatorname{qr}$ returns the orthonormal factor of a Householder QR decomposition.
The resulting $Q\in\mathbb R^{d\times k}$ is used to project the proposed parameter update onto the complement of the protected subspace, $\Delta'=\Delta(I-QQ^\top)$.

\subsection{Hardware and Pipeline Scheduling}

Experiments are performed on a GPU cluster comprising $3$ nodes with $8$ NVIDIA RTX 3090 (24 GB) GPUs per node, with RTX A6000 GPUs also used for training runs.
Each run uses $8$ GPUs, and wall-clock measurements are conducted on a single node with $8$ RTX 3090 GPUs.
Communication uses the \texttt{gloo} backend.

The pipeline depth varies over
$P\in\{1,2,4,8,16,32\}$.
When $P>8$, multiple virtual pipeline stages are assigned to each physical GPU, giving $P/8$ stages per GPU.
This preserves the logical pipeline depth, the number of in-flight micro-batches, and the order of forward, backward, and update operations.
The resulting wall-clock measurements may include additional context-switching overhead, which is shared across the compared methods.

\section{Hyperparameter Ablations}
\label{app:ablation}

We vary the peak learning rate, $\beta_1$, and learning-rate damping at $P=32$, and report the learning rate selected at each depth.
Each point is the lowest evaluation loss over $\beta_1$ (unless varied) and the method-specific setting ($\alpha$ for \ours{}, $D$ for Recovery); all runs use a single seed.

\begin{figure}[t]
    \centering
    \begin{subfigure}{0.24\linewidth}
        \centering
        \includegraphics[width=\linewidth]{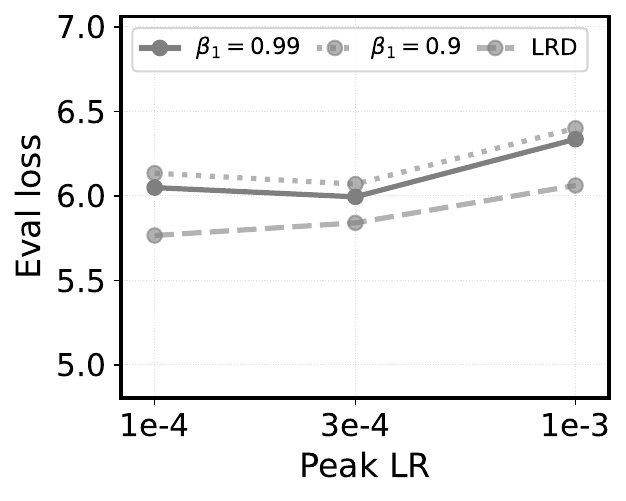}
        \caption{Inconsistent}
    \end{subfigure}
    \hfill
    \begin{subfigure}{0.24\linewidth}
        \centering
        \includegraphics[width=\linewidth]{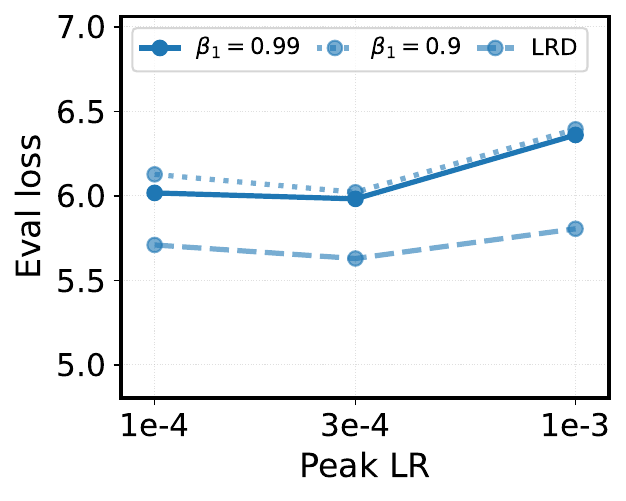}
        \caption{Parameter stashing}
    \end{subfigure}
    \hfill
    \begin{subfigure}{0.24\linewidth}
        \centering
        \includegraphics[width=\linewidth]{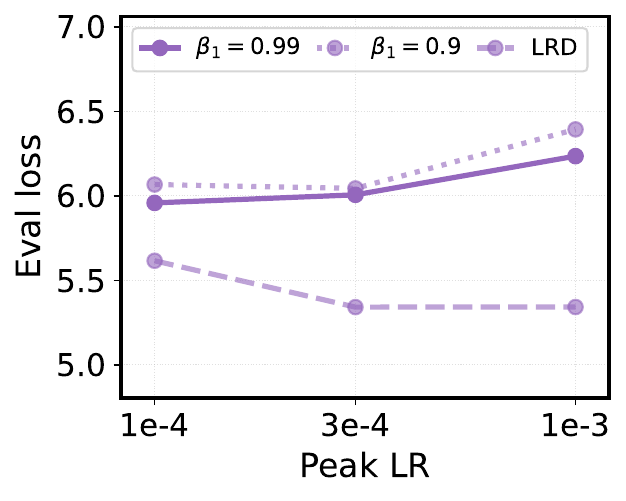}
        \caption{Parameter recovery}
    \end{subfigure}
    \hfill
    \begin{subfigure}{0.24\linewidth}
        \centering
        \includegraphics[width=\linewidth]{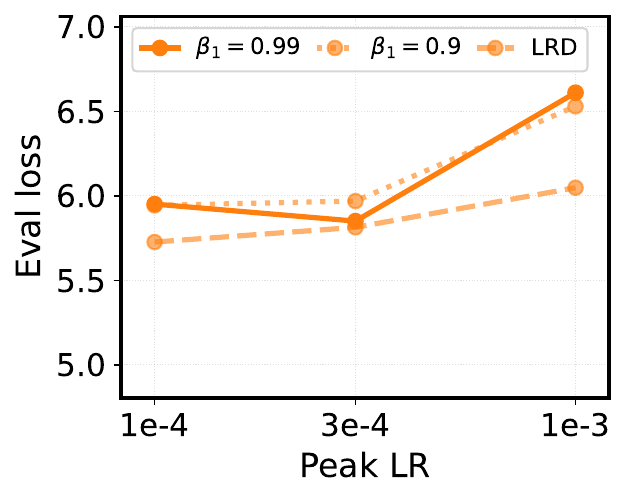}
        \caption{PipeOptim}
    \end{subfigure}

    \vspace{0.8em}

    \begin{subfigure}{0.24\linewidth}
        \centering
        \includegraphics[width=\linewidth]{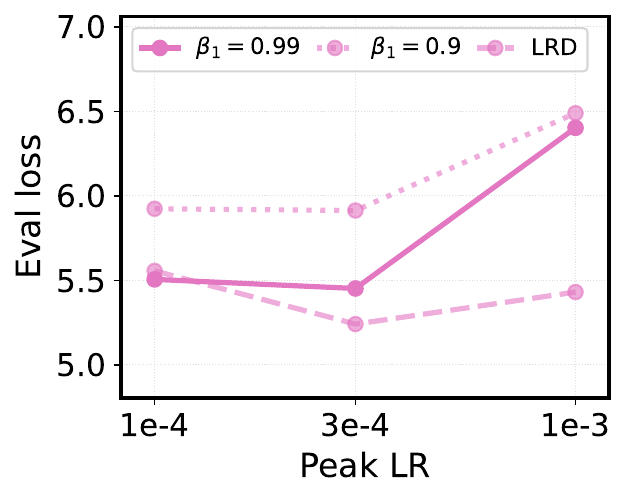}
        \caption{LWP}
    \end{subfigure}
    \hfill
    \begin{subfigure}{0.24\linewidth}
        \centering
        \includegraphics[width=\linewidth]{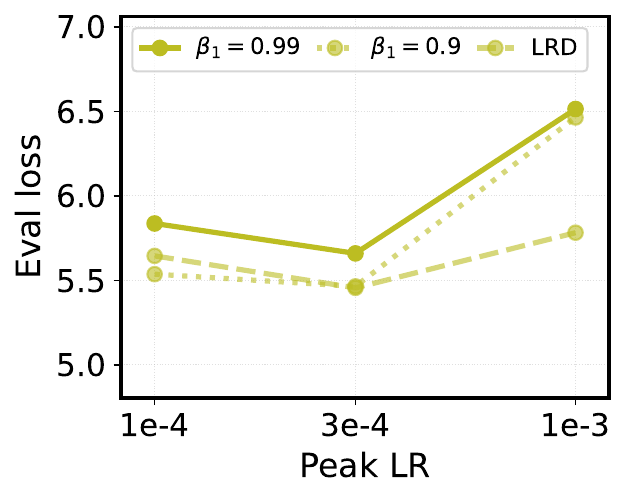}
        \caption{SpecTrain}
    \end{subfigure}
    \hfill
    \begin{subfigure}{0.24\linewidth}
        \centering
        \includegraphics[width=\linewidth]{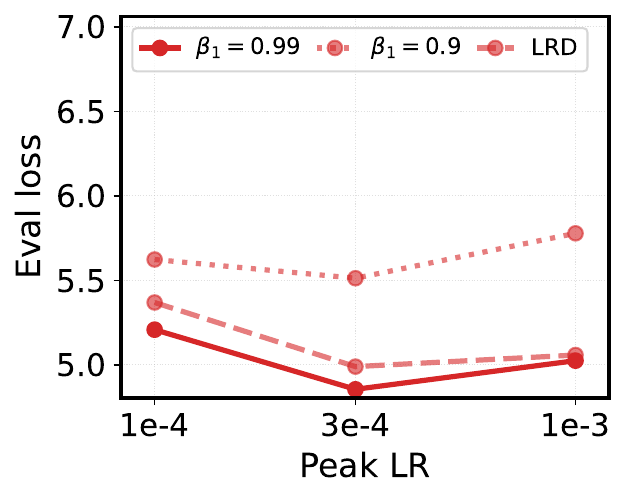}
        \caption{\ours{}}
    \end{subfigure}
    \hfill
    \begin{subfigure}{0.24\linewidth}
        \centering
        \includegraphics[width=\linewidth]{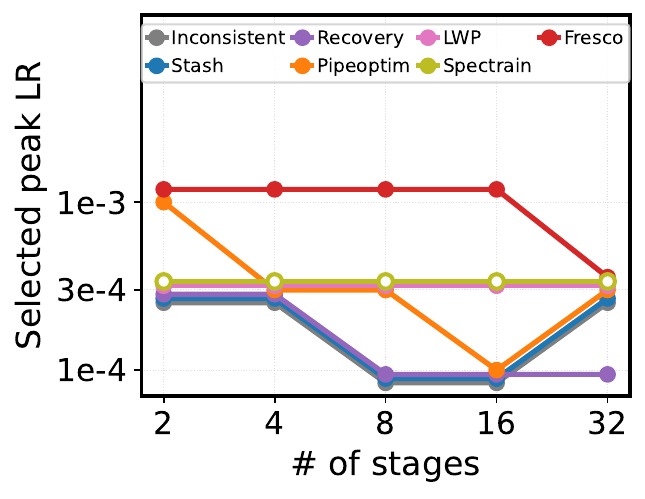}
        \caption{Selected LR per depth}
        \label{fig:lr_depth}
    \end{subfigure}
    \caption{
    Learning rate, momentum, and learning-rate damping.
    (a--g) Evaluation loss over the peak learning rate at $P=32$ with $\beta_1=0.99$ (solid) and $\beta_1=0.9$ (dotted), both without LRD, and with LRD at the best $\beta_1$ (dashed).
    (h) The peak LR selected for each method and pipeline depth in the main results (\cref{fig:main_result_combined}); hollow markers denote $\beta_1=0.9$, and coinciding lines are offset slightly for visibility.
    }
    \label{fig:ablation_lr}
\end{figure}

\subsection{Learning rate}
Every method except Recovery attains its lowest evaluation loss at a peak LR of $3\times10^{-4}$ (\cref{fig:ablation_lr}), and \ours{} has the lowest loss at every learning rate.
\ours{} is also the least sensitive to a larger step size: raising the LR from $3\times10^{-4}$ to $10^{-3}$ increases its loss by $0.17$, compared with $0.23$--$1.00$ for the baselines.
The three prediction-based methods degrade the most (PipeOptim $0.68$, LWP $0.95$, SpecTrain $1.00$), consistent with larger updates making their predicted weights less accurate.

Across depths (\cref{fig:lr_depth}), Inconsistent, Stash, and Recovery select the smallest LR ($10^{-4}$) at $P=8$ and $16$ (Recovery also at $P=32$), and PipeOptim at $P=16$.
\ours{} instead selects the largest LR ($10^{-3}$) up to $P=16$ and $3\times10^{-4}$ at $P=32$, a $10\times$ larger step than these baselines at $P=8$ and $16$.
LWP and SpecTrain select $3\times10^{-4}$ at every depth.

\subsection{Momentum}
$\beta_1=0.99$ gives a lower loss than $\beta_1=0.9$ for every method except SpecTrain ($5.66$ versus $5.47$; \cref{fig:ablation_lr}).
The coefficient matters most for \ours{} ($5.51\to4.86$) and LWP ($5.91\to5.45$); for the other baselines the difference is at most $0.09$.
Across depths, every method that is swept over both values selects $\beta_1=0.99$ (\cref{fig:lr_depth}).
Because the preferred $\beta_1$ differs across methods, we tune it per method rather than fixing it.

\subsection{Learning-rate damping}
We evaluate the staleness-aware learning-rate damping (LRD) of PipeMare~\citep{yang2021pipemare}, which sets the learning rate of a stage with delay $\tau$ to $\eta_t\tau^{-p_t}$ with $p_t=1-\min(t/K,1)$, where $\eta_t$ is the scheduled learning rate at step $t$; we use $K=15{,}000$.
LRD improves every baseline except SpecTrain, which is essentially unaffected ($5.47\to5.46$; \cref{fig:ablation_lr}); the largest gain is for Recovery ($5.96\to5.34$), whose discrepancy correction PipeMare proposes together with LRD.
In contrast, LRD increases the evaluation loss of \ours{} from $4.86$ to $4.99$ and shifts its best budget to the smallest one, $\alpha=12/384$ (\cref{fig:alpha_lrd}).
One possible explanation is that LRD and the projection of \ours{} both attenuate the updates that would interfere with in-flight micro-batches, so that combining them over-damps optimization.
Even when LRD is enabled for every baseline, \ours{} without LRD remains the best configuration, $0.39$ below the strongest damped baseline (LWP, $5.24$), although LRD reorders the baselines (e.g., Recovery moves from fourth to second).
We therefore disable LRD for all methods in the main comparisons, so that every method uses the same learning-rate schedule.

\subsection{Protection budget}

\begin{figure}[!htb]
    \centering
    \begin{subfigure}{0.32\linewidth}
        \centering
        \includegraphics[width=\linewidth]{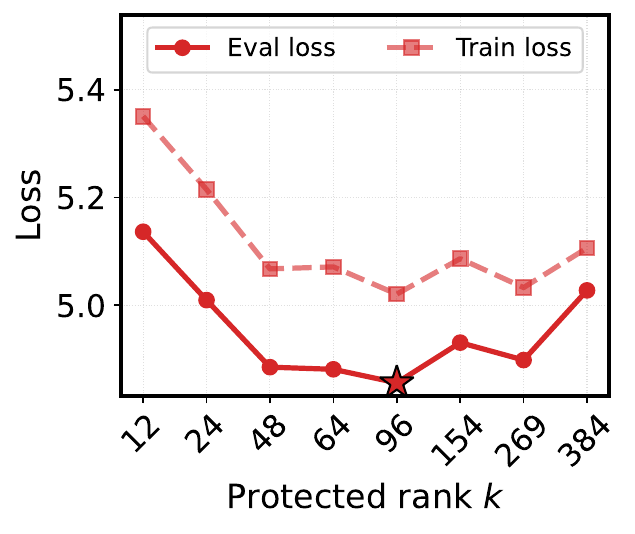}
        \caption{Budget at $P=32$}
        \label{fig:alpha_p32}
    \end{subfigure}
    \hfill
    \begin{subfigure}{0.32\linewidth}
        \centering
        \includegraphics[width=\linewidth]{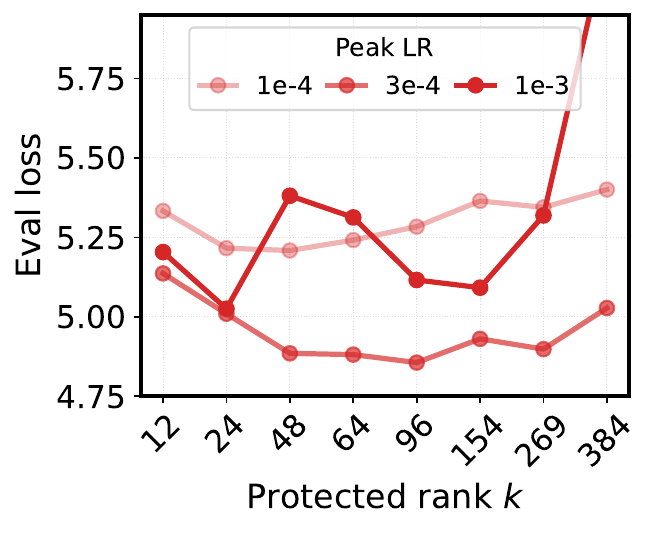}
        \caption{Budget and LR}
        \label{fig:alpha_lr}
    \end{subfigure}
    \hfill
    \begin{subfigure}{0.32\linewidth}
        \centering
        \includegraphics[width=\linewidth]{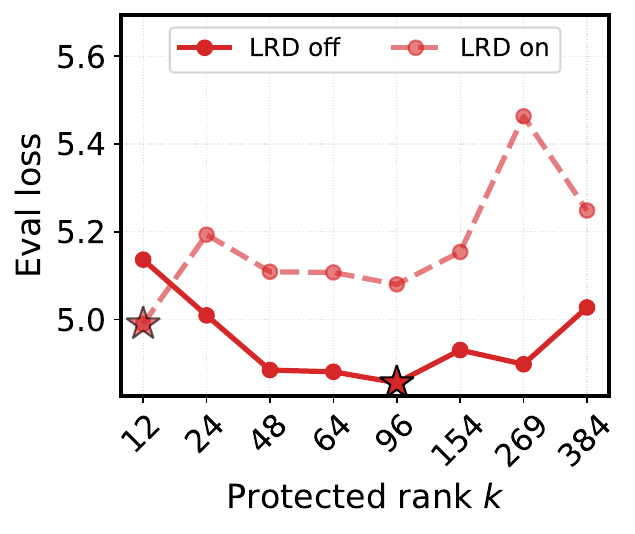}
        \caption{Budget and LRD}
        \label{fig:alpha_lrd}
    \end{subfigure}
    \caption{
    Protection budget of \ours{} over the protected rank $k$ ($\alpha=k/384$).
    (a) Evaluation loss and training loss (EMA $0.97$ at step $19{,}900$) at $P=32$, each budget at its best peak LR and $\beta_1$ by evaluation loss ($3\times10^{-4}$ and $0.99$ for every budget).
    (b) One line per peak LR at $P=32$.
    (c) LRD off and on at $P=32$.
    Stars in (a) and (c) mark the best budget of each line.
    }
    \label{fig:ablation_alpha}
\end{figure}

At $P=32$, the evaluation loss is lowest at $\alpha=96/384$ and stays within $0.08$ of it over $\alpha\in[48/384,269/384]$ (\cref{fig:alpha_p32}).
A peak LR of $3\times10^{-4}$ gives the lowest loss at every budget (\cref{fig:alpha_lr}), so the budget can be selected at a single learning rate.
At $10^{-3}$, however, large budgets degrade sharply, up to an evaluation loss of $6.19$ at full rank, while at $10^{-4}$ the loss varies by less than $0.2$ across budgets.

\end{document}